\pdfoutput=1

\documentclass[11pt]{article}

\usepackage{ACL2023}

\usepackage{times}
\usepackage{latexsym}
\usepackage[T1]{fontenc}
\usepackage[utf8]{inputenc}
\usepackage{microtype}
\usepackage{inconsolata}
\usepackage{hyperref}
\usepackage{times}
\usepackage{latexsym}
\usepackage{hyperref}
\usepackage{latexsym}
\usepackage{amsthm}
\usepackage{amsfonts}
\usepackage{graphicx}
\usepackage{multirow}
\usepackage{pbox}
\usepackage{float}
\usepackage[export]{adjustbox}
\usepackage{tabularx}
\usepackage{graphicx}
\usepackage{longtable}
\usepackage{makecell}
\usepackage{cellspace}
\usepackage{colortbl}
 
\usepackage[font=small,skip=0pt]{caption}
\usepackage[tableposition=below]{caption}
\usepackage{subcaption}
\title{Discourse Relations Classification and Cross-Framework Discourse Relation
Classification Through the Lens of Cognitive Dimensions: An Empirical Investigation}

\author{Yingxue Fu \\
    School of Computer Science \\ University of St Andrews\\ KY16 9SX, UK \\
  \texttt{yf30@st-andrews.ac.uk} \\}

\begin{document}
\maketitle
\begin{abstract}
Existing discourse formalisms use different taxonomies of discourse relations, which require expert knowledge to understand, posing a challenge for annotation and automatic classification. We show that discourse relations can be effectively captured by some simple cognitively inspired dimensions proposed by~\citet{sanders2018unifying}. 
Our experiments on cross-framework discourse relation classification (PDTB \& RST) demonstrate that it is possible to transfer knowledge of discourse relations for one framework to another framework by means of these dimensions, in spite of differences in discourse segmentation of the two frameworks. This manifests the effectiveness of these dimensions in characterizing discourse relations across frameworks. Ablation studies reveal that different dimensions influence different types of discourse relations. The patterns can be explained by the role of dimensions in characterizing and distinguishing different relations. We also report our experimental results on automatic prediction of these dimensions.

\end{abstract}

\section{Introduction}
Discourse relations are useful for various downstream NLP tasks, such as text generation~\citep{ji-huang-2021-discodvt} and machine translation~\citep{sim-smith-2017-integrating}. However, discourse relations are shaped by multiple sources of information and require expert knowledge for annotation. Since the release of the Penn Discourse Treebank 2.0 (PDTB 2.0)~\citep{prasad-etal-2008-penn}, less than 8\% improvement has been made in English implicit relation classification in more than ten years~\citep{atwell-etal-2021-discourse}. Even with the development of contextualized embeddings, this task shows the least improvement in performance compared with other NLP tasks.

Another issue is that existing studies on discourse relation classification are separated into several independent strands of work~\citep{zeldes-etal-2021-disrpt}. The complex nature of discourse gives rise to discourse annotation frameworks which vary in assumptions and definitions of fundamental aspects of discourse, such as what constitutes a discourse relation, what is a basic discourse unit, full-coverage or shallow discourse annotation, and how discourse structure is represented~\citep{fu-2022-towards}.

The leading examples of these annotation frameworks include the Rhetorical Structure Theory (RST)~\citep{mann1988rhetorical}, the Segmented Discourse Representation Theory (SDRT)~\citep{asher2003logics} and the Discourse Lexicalized Tree-Adjoining Grammar (D-LTAG)~\citep{forbes2003d}. These three frameworks have been used in various discourse annotation projects covering different languages. Based on the RST framework, the Rhetorical Structure Theory Discourse Treebank (RST-DT)~\citep{carlson-etal-2001-building} is developed. SDRT forms the theoretical framework for the ANNODIS corpus~\citep{afantenos-etal-2012-empirical}, the STAC corpus~\citep{asher-etal-2016-discourse} and so on, and D-LTAG is the theoretical foundation for PDTB~\citep{prasad-etal-2008-penn,ldcexample}, which is the largest corpus annotated with discourse relations.  

To enable different strands of research to come together and benefit from data across frameworks, we need an interface with which discourse relation classification tasks under different frameworks can be formulated in similar terms, independent of the underlying theoretical assumptions~\citep{zeldes-etal-2021-disrpt}. The UniDim proposal by~\citet{sanders2018unifying} represents one of the influential approaches for this task. The intuition is that discourse relations of different frameworks can be decomposed into cognitive primitives rooted in the Cognitive approach to Coherence Relations (CCR)~\citep{sanders1992toward, sanders1993coherence} (hence denoted as the CCR framework), and people can make use of these elementary notions to relate and compare discourse relations. These primitives are not intended to form a complete and descriptively adequate account of discourse relations but are targeted at a psychologically plausible theory of discourse relations~\citep{sanders1992toward}. Additional primitives are added in later studies to reach better linguistic and cognitive coverage~\citep{crible2019domains}.

~\citet{sanders2018unifying} and other researchers~\citep{rehbein-etal-2016-annotating} try to test if discourse relations annotated based on the CCR framework are consistently categorized into relations under other frameworks. Their investigation reveals that discrepancies between frameworks arise due to variations in how coherence relations are defined, the methods used to perform the annotation, and the rules governing segmentation, and the alignment of discourse relations is generally many-to-many.  

In this study, we aim to assess to what extent these CCR dimensions provide information about discourse relations of different frameworks. We assume that CCR dimensions are annotated in parallel to discourse relation annotations of other frameworks and utilize these dimensions as features in discourse relation classification tasks. The improvement/degradation of performance relative to the case without such features as a measure of the information that these dimensions provide. In this way, we show empirical evidence of the effectiveness of the UniDim proposal in representing and bridging discourse relations of different discourse annotation schemes.  

Our contributions include: 
\begin{itemize}
\item We show that the dimensions of the UniDim proposal~\citep{sanders2018unifying} effectively capture discourse relations and are useful for training computational systems for discourse relation classification, both for RST relation classification and PDTB explicit and implicit relation classification, yielding significant performance gains. Such elementary cognitive dimensions can be useful features for the challenging task of discourse relation classification.       

\item We demonstrate that these dimensions can work as an interface for discourse relations across different frameworks. It is possible to train one discourse relation classification model on PDTB and apply the model to the discourse relation classification task in RST with transfer learning and the performance is as high as training a model specifically for RST relation classification, in spite of differences in discourse segmentation between the two frameworks. The CCR dimensions provide an effective means of bridging discourse relations of different frameworks.  

\item We report experimental results on automatic prediction of these dimensions with RST-DT, PDTB 3.0 and a combination of the two corpora. 
\end{itemize}

\section{Related Work}
\subsection{Mapping Discourse Relations of Different Frameworks}
Prior studies on mapping discourse relations of different frameworks adopt varied approaches. Some researchers propose common inventories of relations that are created based on analysis of discourse relations of different frameworks~\citep{ benamara-taboada-2015-mapping, bunt2016iso}. Alternatively, an intermediate representation may be used to reduce the number of mappings necessary to harmonize different frameworks~\citep{chiarcos-2014-towards, sanders2018unifying}. As there are corpora that contain parallel annotations under different frameworks on the same texts, these corpora are used to identify mappings between discourse relations. Since this approach relies on textual matching, differences in discourse segmentation would hinder relation mapping, leaving only a small number of relations successfully mapped between different frameworks~\citep{bourgonje-zolotarenko-2019-toward, scheffler2016mapping}. The study by~\citet{demberg2019compatible} employs the \textit{strong nuclearity hypothesis}~\citep{marcu2000theory} to mitigate this problem. 
\citet{demberg2019compatible} show that the Unified Dimension (UniDim) approach is relatively successful in mapping relations between RST-DT and PDTB 2.0. 

~\citet{roze-etal-2019-aspects} investigates the possibility of predicting CCR dimensions automatically. They achieve an accuracy above the baseline of majority class guessing. Furthermore, they try to predict relations of PDTB 2.0 from these dimensions, and it is shown that the accuracy is much lower than that of training a model for predicting PDTB relations directly. The low performance may be attributed to the high level of under-specification in the mapping from PDTB relations to these dimensions and the reverse mapping from dimension combinations to the hierarchical PDTB sense labels, especially when the mapping is not necessarily one-to-one. 

Recent studies propose to represent discourse relations as question-answering (QA) pairs~\citep{ko-etal-2022-discourse, pyatkin-etal-2020-qadiscourse}. While this approach is designed to simplify discourse relation labelling, some relations cannot be expressed by QA pairs~\citep{pyatkin-etal-2020-qadiscourse}, and evaluation is difficult. Moreover, open-ended QA leads to annotations similar to the GraphBank~\citep{wolf-gibson-2004-representing}, which has higher complexity than the other frameworks. 

\subsection{Dimensions in UniDim Proposal}
The main approach adopted in the UniDim proposal is to use cognitively inspired dimensions as an intermediate representation and decompose discourse relations of different frameworks into these dimensions so that they can be related and compared. The result contains five dimensions which are rooted in the Cognitive approach to Coherence Relations (CCR)~\citep{sanders1992toward, sanders1993coherence} and some additional dimensions that are added to allow more relations to be better represented (collectively referred to as ``UniDim dimensions'' or ``dimensions of the UniDim proposal'' in the following). We give an overview of these dimensions here.

Two segments that may stand in a discourse relation are identified first, the two segments being denoted as~\textit{$S_1$} and~\textit{$S_2$} in linear order, and the underlying propositions being denoted as~\textit{$P$} and~\textit{$Q$} in linear order. 

The first dimension is \textbf{basic operation}, which has two values:~\textit{causal} and~\textit{additive}. A causal relation means that the two segments are strongly connected and typically, an implication relation $P \rightarrow Q$ can be deduced. In (1),~\textit{$S_2$} shows the cause and~\textit{$S_1$} gives the consequent. If the two segments are just loosely connected and only a conjunction relation $P \land Q$ can be inferred, the value at this dimension is additive, as shown in (2).

(1) [He immigrated to the US,]$_{S_1}$ because [his natural parents were believed to live there.]$_{S_2}$ 

(2) [She is a painter]$_{S_1}$ and [her studio is a few blocks away.]$_{S_2}$ 

As indicated in~\citet{sanders2018unifying}, basic operation can be used to distinguish causal relations or conditional relations from additive relations or temporal relations. 

The second dimension is \textbf{source of coherence}. It has two values: semantic and pragmatic in the original proposal ~\citep{sanders1992toward}, later renamed as~\textit{objective} and~\textit{subjective} in~\citet{maat2000domains}, respectively. A relation is objective if the segments are connected because of their propositional content, and the relation holds because the connection is coherent based on world knowledge, as shown in (3). A relation is subjective if the speaker's reasoning or the pragmatic effect of the relation is prominent. (4) shows a claim in~\textit{$S_2$} and ~\textit{$S_1$} is an argument that supports it. 

(3) [It was dark outside,]$_{S_1}$ so [he lit up a candle.]$_{S_2}$ 

(4) [Smoking is unhealthy]$_{S_1}$ and [we should put a limit on it.]$_{S_2}$ 

This dimension can be used to distinguish relations that are related to real-world situations, such as temporal sequence, and cause-consequence, from argumentative relations, such as claim-argument or evidence-justification~\citep{sanders2018unifying}. 

The third dimension is \textbf{implication order}. This dimension distinguishes between~\textit{non-basic} and~\textit{basic} orders of causal relations, and does not apply to additive relations, which are generally symmetric. For a causal relation characterized by $P \rightarrow Q$, if~\textit{$S_1$} expresses~\textit{$P$} and~\textit{$S_2$} expresses~\textit{$Q$} (note that~\textit{$S_1$} and~\textit{$S_2$} are in linear order), then this relation is in basic order, as shown in (6). If~\textit{$S_2$} actually expresses~\textit{$P$} while ~\textit{$S_1$} expresses~\textit{$Q$}, this relation is in non-basic order, as shown in (5).

(5) [He did not attend the conference,]$_{S_1}$ because [he received a message telling him not to go.]$_{S_2}$ 

(6) Because [he received a warning message,]$_{S_1}$ [he did not attend the conference.]$_{S_2}$ 

It is clear to see that the implication order dimension is mainly used to distinguish relations with directionality, such as cause-result and cause-reason. 

The fourth dimension is \textbf{polarity}. A relation is characterized by~\textit{positive} polarity if the propositions~\textit{$P$} and \textit{$Q$}, expressed by~\textit{$S_1$} and~\textit{$S_2$}, respectively, have the same logical polarity and support each other, as shown in (7). A relation is of~\textit{negative} polarity if the relation involves the juxtaposition of $\neg$\textit{$P$} and \textit{$P$} or $\neg$\textit{$Q$} and \textit{$Q$} in the two segments, as shown in (8). In this example, a positive polarity would require a reason or result that supports the decision of closing the library.

(7) [We like the garden]$_{S_1}$ because [it is pretty.]$_{S_2}$ 

(8) [The university library was closed]$_{S_1}$ although [students wanted more space for study.]$_{S_2}$  

This dimension is useful for capturing contrastive, adversative and concession relations~\citep{sanders2018unifying}.

The fifth dimension is \textbf{temporality}, which distinguishes between~\textit{temporal} and~\textit{non-temporal} relations. Under temporal relations, temporality has three values:~\textit{synchronous},~\textit{chronological} and~\textit{anti-chronological}. Synchronous relations are those temporal relations which feature simultaneous occurrence of events. If events described in the segments happen in temporal order, then the relation is chronological, otherwise the relation is anti-chronological. 

In order to characterize more relations, additional dimensions are introduced, including~\textbf{specificity},~\textbf{lists} and~\textbf{alternatives} for additive relations, and~\textbf{conditionals} and~\textbf{goal-oriented relations} for causal relations (denoted collectively as ``additional dimensions'' in the following). 

\section{Methodology}
Since RST-DT and PDTB both use the WSJ articles of the Penn Treebank, cross-framework relation classification of RST and PDTB by automatic means would be less influenced by domain shift. Therefore, we focus on the two frameworks. For PDTB, we use PDTB 3.0, which is newer and introduced systematic changes.

As we are primarily interested in the effectiveness of UniDim dimensions rather than improving algorithms for discourse relation classification, simple models are implemented in the experiments.  

\subsection{Discourse Relation Classification} \label{section:3.1}
Discourse relation classification is a typical multi-class classification task. Given a span/argument pair with tokens $S=$ $[CLS]$, $S^{(1)}_1$ $...$ $S^{(1)}_m$, $[SEP]$, $S^{(2)}_1$ $...$ $S^{(2)}_n$, we obtain the representation of the sequence from a pre-trained language model, denoted as $f_{PLM} (S)$, and the embeddings of the dimensions $E$ are obtained from embedding layers, where the embeddings are initialized from uniform distributions and trainable. The representation of the input and the embeddings of dimensions are concatenated:
\begin{equation}\label{input-rep}
h_{S} = f_{PLM}(S) \oplus E{dim_{pol}} \oplus E{dim_{bop}} \oplus ...
\end{equation}
The $dim_{pol}$ and $dim_{bop}$ ... represents the UniDim dimensions, including polarity, basic operation, implication order, source of coherence, temporaltiy, specificity, alternative, conditional and goal.

The representation is fed to two two-layer feed-forward networks (FFNs) with LeakyReLU as activation functions:
\begin{equation}
\hat{h} = g_2 (W_{2} * g_1 ( W_{1}* h_{S} ))  
\end{equation}
where $g_1$ and $g_2$ represent the non-linear activation functions of first and the second FFNs, respectively. $W_1$ and $W_2$ denote weights of the first layers of the two FFNs, and bias terms are omitted for clarity.

A classifier layer is configured on top of the second FFN. 
The predicted result $\hat{y}$ is obtained with:
\begin{equation}
\hat{y} = softmax (W_{3}*\hat{h})  
\end{equation}

Cross-entropy loss is used in the loss function:
\begin{equation} \label{ce-loss} 
 \mathcal{L}_{c}=-\sum\limits_{i=1}^{N}\sum\limits_{l=1}^{C} c^{i}_l
 \log p(c^{i}_l)         
\end{equation}
where $N$ is the batch size, $C$ is the total number of classes, and $p(c^{i}_l)$ is the probability predicted for a class $c$. 

In this design, we take our experiments with transfer learning for cross-framework discourse relation classification into consideration, as we try to keep the architecture and only replace the last classifier layer to fit the model on new data. Moreover, our preliminary experiments indicate that removing the second FFN causes a significant performance drop.

\textbf{Baseline model} The BertForSequenceClassification model from the Transformers library~\citep{wolf-etal-2020-transformers} is used as the baseline model, in which a classifier layer is added on top of the contextualized embeddings of the input sequence.
For an input sequence $S$,  
its representation is obtained with:
\begin{equation} \label{rep-from-lm}
h_{S} = f_{PLM}({S})       
\end{equation}

The predicted result $\hat{y}$ is obtained with:
\begin{equation}
\hat{y} = softmax( W_b* h_{S})    
\end{equation}


As shown in~\citet{kim-etal-2020-implicit}, this model is a strong baseline. We use the \textit{bert-base-uncased} BERT model in all our experiments for comparison of experimental results.

\subsection{Cross-framework Discourse Relation Classification}\label{section:3.2}
We hypothesize that if UniDim dimensions form an effective ``interlingua'' of discourse relations from different frameworks, we can train a model for discourse relation classification in one framework and apply the model for relation classification in another framework without much modification. The transfer learning framework can be used for this experiment. 

As PDTB 3.0 is much larger than RST-DT, a natural choice would be to treat PDTB relation classification as the source task and RST relation classification as the target task~\citep{wang2019characterizing}.   


We first train a model as described in section~\ref{section:3.1} on all the PDTB data, and freeze all the layers but the last classifier layer so that the model can be fit on RST data. 

Formally, for a pair of PDTB arguments $P=$ $[CLS]$, $A^{(1)}_1$ $...$ $A^{(1)}_m$, $[SEP]$, $A^{(2)}_1$ $...$ $A^{(2)}_n$, we obtain the representation of sequence $P$ with equation (1). Through training, the parameters in equation (2) are learnt for the PDTB relation classification task. With these parameters, for an RST span pair $R = [CLS], R^{(1)}_1,..., R^{(1)}_m, [SEP], R^{(2)}_1,..., R^{(2)}_n$, we first obtain the representation of sequence $R$ with equation (1), denoted as $h_{R}$, and with the parameters learnt for PDTB relation classification, we obtain the representation $\widehat{h_{R}}$:
\begin{equation}
\widehat{h_{R}} = g_2 (W_{2} * g_1 ( W_{1}* h_{R}))
\end{equation}

The predicted result $\hat{y}$ for RST relation classification is obtained with:
\begin{equation}
\hat{y} = softmax( W_r* \widehat{h_{R}})
\end{equation}
where $W_r$ is the weight to be learnt for RST relation classification. 

\textbf{Baseline model} As we transfer knowledge from PDTB relation classification to RST relation classification, the baseline model is a model trained specifically for RST relation classification with BERT embeddings and UniDim dimensions as input. For the baseline model in section~\ref{section:3.1}, where only BERT embeddings are used, we train a model for PDTB relation classification and apply the model to RST relation classification without using UniDim dimensions. 

\subsection{Automatic UniDim Dimension Prediction}
Since the dimensions may be related to each other, we train one model for predicting the nine dimensions in equation \ref{input-rep} together.

For an input sequence $S$, 
we obtain its representation $h_{S}$ with equation \ref{rep-from-lm}. A two-layer FFN $f$ with LeakyReLU activation function is applied to $h_{S}$ before nine classification layers $c_{i|i=1...9}$ are applied: 
\begin{equation}
\hat{y} = softmax( W_{c_i} * f (h_{S}) )
\end{equation}
We train the model on PDTB, RST and the combination of PDTB and RST data, respectively. The results reported in~\citet{roze-etal-2019-aspects} are our baseline.

\section{Experiments}
We use the mapping table given in~\citet{sanders2018unifying} (Appendix~\ref{rst-to-unidim-mapping-table}) for obtaining the dimension values for relation labels of RST-DT. As no mapping table is provided for PDTB 3.0, we create the mapping table by ourselves (Appendix~\ref{pdtb-to-unidim-mapping-table}). 

\subsection{Data Preprocessing}

We binarize the RST trees based on the procedure in~\citet{ji-eisenstein-2014-representation} and extract pairs of spans that are connected by a relation. Following~\citet{sanders2018unifying}, we exclude \textit{Same-Unit} and \textit{Attribution} relations from RST-DT, leaving 16 relations. We use the standard split of the corpus and take 20\% from the training set for validation. 

Since PDTB level-2 relations carry specific and generally more useful information, we focus on level-2 relation classification for PDTB. We exclude relations that have fewer than 100 instances to alleviate data imbalance, as suggested in~\citet{kim-etal-2020-implicit}. We follow the data split in~\citet{ji-eisenstein-2015-one}, using sections 2-20 for training, 0-1 for validation and 21-22 for testing. 

We use the pre-trained BERT model~\citep{devlin-etal-2019-bert} for obtaining contextualized embeddings and the [CLS] and
[SEP] tokens are inserted following the settings of the BERT model, which is shown to benefit inter-sentential~\citep{shi-demberg-2019-next} and intra-sentential~\citep{zhao-webber-2021-revisiting} implicit discourse relation classification.

Among the UniDim dimensions, we exclude \textit{list} because this dimension is proposed for representing the \textit{List} relation in PDTB, which has been removed from the sense hierarchy in PDTB 3.0. Following~\citet{roze-etal-2019-aspects}, we merge \textit{specificity-example} and \textit{specificity-equivalence} into specificity, and add the \textit{NS} label in cases of ambiguity or under-specification. The \textit{N.A.} label is kept when it appears on its own to reflect the fact that some dimensions do not apply to certain types of relations. The default values of additional dimensions are set to negative because they are only applicable to some relations and typically have binary values.   
 
 On the whole, the dimensions are heavily imbalanced and have high degree of under-specification. Statistics for the distribution of these dimensions are shown in Appendix~\ref{unidim-distributions}. Hyper-parameter settings and model training details are described in Appendix~\ref{hyperparameter}.

\subsection{Evaluation}
For RST relation classification, the settings of the DISRPT 2021 shared task on relation classification~\citep{zeldes-etal-2021-disrpt} are the closest to ours. We report their best accuracy on RST-DT~\citep{gessler-etal-2021-discodisco} alongside our baseline model results for comparison. 

After preprocessing, we perform 12-way explicit relation classification and 14-way implicit relation classification for PDTB. While most of the previous studies use PDTB 2.0 and recent studies on PDTB 3.0 only focus on implicit relation classification, when settings of previous studies are close to ours, we report their results alongside our baseline results\footnote{We build and run all the baseline models mentioned in section~\ref{section:3.1} and section~\ref{section:3.2} by ourselves.}.       


\subsection{Results and Discussion}
We report our experimental results on the test sets, which are computed with the Scikit-Learn library~\citep{scikit-learn}. We can expect that RST and PDTB data show different patterns. For RST, the dimension values for end labels may be clear, but when end labels are grouped into a class, the values could be rather mixed. For PDTB, as L2 sense classification is performed, the process of grouping relations into broader classes happens at L3, which only encodes directionality, and dimensions that are related to directionality are affected, such as implication order, but the other dimensions are not influenced. Therefore, dimension values for PDTB classes tend to be less ambiguous. Moreover, data amount differences are likely to have notable influence on the results. We do not report the results of additional dimensions separately because their individual effects are not obvious.

\subsubsection{RST Relation Classification}
Table~\ref{body-rst-cls-dim-report} shows results on RST-DT. 
\begin{table}[h!]\tiny
\centering
\resizebox{0.85\linewidth}{!}{%
\begin{tabular} {|p{0.9cm}| >{\columncolor[RGB]{230, 242, 255}}c|>{\columncolor[RGB]{230, 242, 255}}c|>{\columncolor[RGB]{230, 242, 255}}c| c|c|c|c|}
\hline  
&\tiny $P$
&\tiny $R$
&\tiny $F1$
&\tiny $P_{b.}$
&\tiny $R_{b.}$
&\tiny $F1_{b.}$
&\tiny $C.$\\ 
\hline
 Background & 1.00 & 1.00 & 1.00 & 0.47 & 0.35 & 0.40 & 111 \\  \hline
Cause & 0.92 & 0.70 & 0.79 & 0.50 & 0.17 & 0.25 & 82  \\ \hline
Comparison & 0.00 & 0.00 & 0.00 & 0.61 & 0.38 & 0.47 &29 \\ \hline
Condition & 1.00 & 1.00 & 1.00 & 0.79 & 0.71 & 0.75 & 48 \\ \hline
Contrast & 0.99 & 1.00 & 0.99 &  0.75 & 0.68 & 0.72 & 146 \\ \hline
Elaboration & 0.75 & 1.00 & 0.86 & 0.65 & 0.88 & 0.75 & 796 \\ \hline
Enablement & 0.92 & 1.00 & 0.96 & 0.61 & 0.85 & 0.71 & 46 \\ \hline
Evaluation & 0.99 & 1.00 & 0.99 & 0.29 & 0.14 & 0.19 & 80 \\ \hline
Explanation & 0.72 & 0.97 & 0.83 &  0.46 & 0.27 & 0.34 & 110 \\ \hline
Joint & 1.00 & 0.03 & 0.06 & 0.67 & 0.62 & 0.64 & 212  \\ \hline
Manner-Means & 0.00 & 0.00 & 0.00 & 0.68 & 0.48 & 0.57 & 27 \\ \hline
Summary & 0.00 & 0.00 & 0.00 & 0.88 & 0.47 & 0.61 & 32 \\ \hline
Temporal & 1.00 & 1.00 & 1.00 & 0.74 & 0.27 & 0.40 & 73  \\ \hline
Textual-Organization & 0.00 & 0.00 & 0.00 & 0.44 & 0.44 & 0.44 & 9 \\ \hline
Topic-Change & 0.28 & 1.00 & 0.44 &  0.28 & 0.38 & 0.32 & 13\\ \hline
Topic-Comment & 0.71 & 0.21 & 0.32 & 0.00 & 0.00 & 0.00 & 24 \\ \hline
\textbf{Acc.}& \multicolumn{3}{c|} {\textbf{0.81}} & \multicolumn{4}{c|} {0.63  (vs DISRPT 2021: 0.67)} \\ \hline
\textbf{Macro-F1} & \textbf{0.64} & \textbf{0.62} & \textbf{0.58} & 0.55 & 0.44 & 0.47 & 1838\\  \hline
\end{tabular} %
}
\vspace{2mm}
\caption{\label{body-rst-cls-dim-report} Results of RST relation classification. The columns in blue show the results of our method and uncolored columns show the results of the baseline model, and the last column shows the count of occurrences of each relation in the test set. We use this convention in reporting the results. }
\end{table}
When UniDim dimensions are added as features, a significant performance gain can be obtained. Some relations can be recognized with 100\% accuracy. However, relations including \textit{Comparison}, \textit{Manner-means}, \textit{Summary} and \textit{Textual-Organization} cannot be recognized. From Fig.\ref{rst-train-stats} in Appendix~\ref{rel-train-stats}, it is clear that these relations have small amounts of training data. As we focus on broader classes rather than end labels in relation classification, we can see from the mapping table in Appendix~\ref{rst-to-unidim-mapping-table} that dimension values under these classes are mixed. It is difficult for the model to learn patterns from the data.    

To have a better understanding of the influence of each dimension on the results, we performed ablation studies and the results are shown in Table~\ref{body-rst-ablation-dim-report}.

\begin{table}[H]\tiny
\centering
\begin{tabular}{|c|c|c|c|c|}
\hline  
&\tiny $Acc$
&\tiny $P$ 
&\tiny $R$
&\tiny $F1$ \\
\hline
Total& 0.81 & 0.64 & 0.62 & 0.58 \\ \hline
-Pol. & 0.74 & 0.49 & 0.48 & \underline{\textbf{0.48}} \\ \hline
-Basic Op. & 0.78 & 0.52 & 0.58 & 0.53 \\ \hline
-SoC. & 0.78 & 0.52 & 0.58 & 0.53 \\ \hline
-Impl. order & 0.81 & 0.58 & 0.60 & 0.55 \\ \hline
-Temp. & 0.80 & 0.59 & 0.60 & 0.55 \\ \hline
-Add. & 0.80 & 0.52 & 0.59 & 0.54 \\ \hline
\end{tabular}
\vspace{2mm}
\caption{\label{body-rst-ablation-dim-report} Results of ablation studies for RST relation classification, showing the overall accuracy ($Acc$), precision ($P$), recall($R$) and macro-averaged F1 ($F1$) for dimensions of polarity (Pol.), basic operation (Basic Op.), source of coherence (SoC.), implication order (Impl. order), temporality (Temp.) and additional dimensions (Add.), respectively. }
\end{table}

As shown in Table~\ref{body-rst-ablation-dim-report}, removing the polarity dimension causes the biggest performance drop in macro-averaged F1. By examining the detailed results (Table~\ref{rst-pol-ab}, Appendix~\ref{rst-rel-cls-ablation-studies}), we find that removing this dimension has noticeable influence on the recognition of \textit{Contrast}($\downarrow$ 0.41), \textit{Evaluation}($\downarrow$ 0.26), \textit{Topic-Change}($\downarrow$ 0.44) and \textit{Topic-Comment}($\downarrow$ 0.32). The correlation between \textit{Contrast} and this dimension is self-evident. Examination of the mapping table suggests that the rest of these relations have ambiguous or mixed values in the other dimensions and their data amounts are small, making it difficult for the model to learn any patterns.  

\subsubsection{PDTB Explicit Relation Classification}
Table~\ref{body-pdtb-expl-cls-dim-report} shows the results of 12-way explicit relation classification. The overall accuracy score is high and the majority of the relations can be recognized with near perfect performance, which means that the UniDim dimensions are effective in characterizing most of the PDTB explicit relations. However, in spite of the noticeable improvement in overall accuracy, our method does not show improvement over the baseline model in macro-averaged F1 score. This is likely due to the strong reliance of pre-trained language models on lexical cues in discourse relation classification tasks~\citep{kim-etal-2020-implicit} and these lexical cues are effective features for this task. Moreover, with our approach, the \textit{Level-of-detail} and \textit{Substitution} relations cannot be recognized. The two relations have the smallest data amount, and in terms of dimension values, \textit{Substitution} is similar to \textit{Concession} and \textit{Level-of-detail} is similar to \textit{Manner}. It is possible that the model predicts \textit{Manner} for instances of \textit{Level-of-detail}, which explains the lower precision for \textit{Manner}.     

\begin{table}[h!]\tiny
\centering
\resizebox{0.85\linewidth}{!}{%
\begin{tabular} {|c| >{\columncolor[RGB]{230, 242, 255}}c|>{\columncolor[RGB]{230, 242, 255}}c|>{\columncolor[RGB]{230, 242, 255}}c| c|c|c|c|}
\hline  
&\tiny $P$
&\tiny $R$
&\tiny $F1$
&\tiny $P_{b.}$
&\tiny $R_{b.}$
&\tiny $F1_{b.}$
&\tiny $C.$\\ 
\hline
 
Asynchronous & 1.00 & 1.00 & 1.00 &  0.97 & 0.87 & 0.92 & 127\\ \hline
Cause & 1.00 & 1.00 & 1.00 & 0.82 & 0.89 & 0.85 & 115 \\ \hline
Concession & 0.96 & 1.00 & 0.98 &  0.89 & 0.95 & 0.92 &  285 \\ \hline
Condition & 1.00 & 1.00 & 1.00 & 0.93 & 0.92 & 0.93 & 61 \\ \hline
Conjunction & 1.00 & 1.00 & 1.00 & 0.97 & 0.96 & 0.96 & 516 \\ \hline
Contrast & 1.00 & 1.00 & 1.00 & 0.52 & 0.48 & 0.50 &  50 \\ \hline
Disjunction & 1.00 & 1.00 & 1.00 &  0.90 & 1.00 & 0.95 & 18 \\ \hline
Level-of-detail & 0.00 & 0.00 & 0.00 & 0.71 & 0.75 & 0.73 & 20 \\ \hline
Manner & 0.35 & 1.00 & 0.52 & 0.42 & 0.91 & 0.57 & 11 \\ \hline
Purpose & 1.00 & 1.00 & 1.00 & 0.62 & 0.45 & 0.52 & 29 \\ \hline
Substitution & 0.00 & 0.00 & 0.00 & 1.00 & 0.92 & 0.96 & 13 \\ \hline
Synchronous & 1.00 & 1.00 & 1.00 &  0.81 & 0.71 & 0.76 & 126 \\ \hline

\textbf{Acc.}& \multicolumn{3}{c|} {\textbf{0.98}} & \multicolumn{4}{c|} {0.89} \\ \hline
\textbf{Macro-F1} & \textbf{0.78} & \textbf{0.83} & \textbf{0.79} & 0.80 & 0.82 & 0.80 & 1371\\ 
\hline
\end{tabular} %
}
\vspace{2mm}
\caption{\label{body-pdtb-expl-cls-dim-report} Results of PDTB explicit relation classification.}
\end{table}


The results of ablation studies are shown in Table~\ref{body-pdtb-expl-ablation-dim-report}. Removing the source of coherence dimension causes the biggest performance drop in macro-averaged F1. Through examining the detailed results, we find that without this dimension, the \textit{Disjunction} relation cannot be recognized. Meanwhile, removing this dimension causes a drop of 0.15 for identifying the \textit{Contrast} relation and a drop of 0.14 for recognizing the \textit{Synchronous} relation. The \textit{Disjunction} relation has a small data amount, and the model might predict \textit{Contrast} for instances of \textit{Disjunction}, since they are similar in the absence of this dimension, which may account for the lower precision for \textit{Contrast}. 

\begin{table}[H]\tiny
\centering
\begin{tabular}{|c|c|c|c|c|}
\hline  
&\tiny $Acc$
&\tiny $P$ 
&\tiny $R$
&\tiny $F1$ \\
\hline
Total& 0.98 & 0.78 & 0.83 & 0.79 \\ \hline
-Pol. & 0.95 & 0.74 & 0.81 & 0.76 \\ \hline
-Basic Op. & 0.98 & 0.78 & 0.83 & 0.79 \\ \hline
-SoC. & 0.94 & 0.67 & 0.73 & \underline{\textbf{0.68}}\\ \hline
-Impl. order & 0.98 & 0.78 & 0.83 & 0.79 \\ \hline
-Temp. & 0.95 & 0.76 & 0.81 & 0.77 \\ \hline
-Add. & 0.96 & 0.73 & 0.73 & 0.73 \\ \hline
\end{tabular}
\vspace{2mm}
\caption{\label{body-pdtb-expl-ablation-dim-report} Results of ablation studies for PDTB explicit relation classification.}
\end{table}

\subsubsection{PDTB Implicit Relation Classification}

Table~\ref{body-pdtb-impl-cls-dim-report} shows the results of 14-way implicit relation classification. The previous best result under similar settings is 0.64 in overall accuracy~\citep{kim-etal-2020-implicit}, which is achieved with \textit{large-cased} XLNet~\citep{yang2019xlnet}. Our baseline 56\% accuracy is consistent with the results in~\citet{kim-etal-2020-implicit}.
\begin{table}[h!]\tiny
\centering
\resizebox{0.85\linewidth}{!}{%
\begin{tabular} {|c| >{\columncolor[RGB]{230, 242, 255}}c|>{\columncolor[RGB]{230, 242, 255}}c|>{\columncolor[RGB]{230, 242, 255}}c| c|c|c|c|}
\hline  
&\tiny $P$
&\tiny $R$
&\tiny $F1$
&\tiny $P_{b.}$
&\tiny $R_{b.}$
&\tiny $F1_{b.}$
&\tiny $C.$\\ 
\hline

Asynchronous & 1.00 & 1.00 & 1.00 & 0.62 & 0.61 & 0.62 & 95\\ \hline
Cause & 1.00 & 1.00 & 1.00 & 0.60 & 0.63 & 0.61 &  366\\ \hline
Cause+Belief & 1.00 & 0.42 & 0.59 &  0.00 & 0.00 & 0.00 &  12 \\ \hline
Concession & 1.00 & 0.92 & 0.96 & 0.44 & 0.40 & 0.42 & 84 \\ \hline
Condition & 1.00 & 1.00 & 1.00 & 0.71 & 0.42 & 0.53 & 12 \\ \hline
Conjunction & 0.90 & 1.00 & 0.95 &  0.49 & 0.61 & 0.54 & 221 \\ \hline
Contrast & 0.98 & 1.00 & 0.99 &  0.45 & 0.42 & 0.43 &  50 \\ \hline
Equivalence & 0.00 & 0.00 & 0.00 & 0.12 & 0.04 & 0.06 &  24 \\ \hline
Instantiation & 0.00 & 0.00 & 0.00 & 0.77 & 0.54 & 0.64 & 107 \\ \hline
Level-of-detail & 0.60 & 1.00 & 0.75 &  0.45 & 0.48 & 0.46 & 180 \\ \hline
Manner & 0.00 & 0.00 & 0.00 & 0.38 & 0.60 & 0.46 & 15\\ \hline
Purpose & 0.92 & 0.94 & 0.93 &  0.92 & 0.98 & 0.95 &  88\\ \hline
Substitution & 0.75 & 1.00 & 0.86 &  0.43 & 0.48 & 0.45 & 21 \\ \hline
Synchronous & 0.87 & 0.97 & 0.92 & 0.27 & 0.10 & 0.15 &  40  \\ \hline

\textbf{Acc.}& \multicolumn{3}{c|} {\textbf{0.87}} & \multicolumn{4}{c|} {0.56} \\ \hline
\textbf{Macro-F1} & \textbf{0.72} & \textbf{0.73} & \textbf{0.71} & 0.48 & 0.45 & 0.45 & 1315\\ 
\hline
\end{tabular} %
}
\vspace{2mm}
\caption{\label{body-pdtb-impl-cls-dim-report} Results of PDTB implicit relation classification.}
\end{table}

As is shown in Table~\ref{body-pdtb-impl-cls-dim-report}, adding UniDim dimensions brings significant performance gain for this task, which is challenging for the baseline model. Meanwhile, we notice that relations including \textit{Equivalence}, \textit{Instantiation} and \textit{Manner} are difficult to recognize. In terms of dimension values, \textit{Equivalence} is similar to \textit{Conjunction}, which has a much larger amount of data. It is likely that the model predicts \textit{Conjunction} for \textit{Equivalence}, hence the lower precision for \textit{Conjunction}. \textit{Instantiation}, \textit{Manner} and \textit{Level-of-detail} have the same dimension values, and as the data amount for \textit{Level-of-detail} is much larger, the model may predict \textit{Level-of-detail} for instances of the other two relations, causing the precision score for \textit{Level-of-detail} to go down.     

The results of ablation studies are shown in Table~\ref{body-pdtb-impl-ablation-dim-report}. Both the implication order dimension and the additional dimensions have substantial influence on the F1 score. Removing the implication order dimension does not cause much decrease in the overall accuracy score but mainly lowers the F1 score, while removing the additional dimensions reduces both the overall accuracy score and the F1 score. 

\begin{table}[H]\tiny
\centering
\begin{tabular}{|c|c|c|c|c|}
\hline  
&\tiny $Acc$
&\tiny $P$ 
&\tiny $R$
&\tiny $F1$ \\
\hline
Total& 0.87 & 0.72 & 0.73 & 0.71 \\ \hline
-Pol. & 0.87 & 0.71 & 0.71 & 0.70 \\ \hline
-Basic Op. & 0.87 & 0.72 & 0.73 & 0.71 \\ \hline
-SoC. & 0.87 & 0.72 & 0.73 & 0.71 \\ \hline
-Impl. order & 0.86 & 0.57 & 0.64 & \underline{\textbf{0.60}} \\ \hline
-Temp. & 0.87 & 0.72 & 0.73 & 0.71 \\ \hline
-Add. & 0.73 & 0.64 & 0.64 & \underline{\textbf{0.62}} \\ \hline
\end{tabular}
\vspace{2mm}
\caption{\label{body-pdtb-impl-ablation-dim-report} Results of ablation studies for PDTB implicit relation classification.}
\end{table}

Detailed results (Table~\ref{impl-pdtb-impl-order-ab} in Appendix \ref{pdtb-impl-cls-ablation-studies}) show that removing the implication order dimension causes a drop of 0.07 in recognizing \textit{Concession}, a drop of 0.86 in recognizing \textit{Substitution} and a drop of 0.59 in recognizing \textit{Cause+Belief}. As the last two relations cannot be recognized, the macro-averaged F1 shows a significant decrease. Similarly, this is associated with differences in data amount and how different relations can be distinguished from each other without the dimension, for instance, \textit{Substitution} has a small data amount, and without the implication order dimension, the model might confuse this relation with \textit{Concession} and predict \textit{Concession} for instances of both relations, which may explain the lower precision for \textit{Concession}. If the additional dimensions are removed, major relations that are impacted include \textit{Condition}($\downarrow$ 0.14), \textit{Conjunction}($\downarrow$ 0.37), and \textit{Level-of-detail}($\downarrow$ 0.75). In this case, the \textit{Level-of-detail} relation cannot be identified. Without this dimension, \textit{Level-of-detail} has the same dimension values as \textit{Conjunction}, which has a larger data amount. The model may predict \textit{Conjunction} for both classes, which causes precision for \textit{Conjunction} to decrease.

\subsubsection{Cross-Framework Discourse Relation Classification}
As RST does not distinguish explicit and implicit relations, we train a model on the whole PDTB data for the source task. We show the overall performance of transfer learning from PDTB to RST in Table~\ref{transfer-learning-pdtb2rst}. The settings of the DISRPT 2021 shared task are the closest to our experiments, and their best results~\citep{gessler-etal-2021-discodisco} are shown alongside the baseline model for comparison. As is clear from the table, the results of transfer learning based on the baseline BERT model show noticeable effect of negative transfer (0.63 $\rightarrow$ 0.58 in overall accuracy and 0.47 $\rightarrow$ 0.33 in F1 score), while with our method, the overall accuracy does not show any decrease and the F1 score is only 1\% lower. This shows that the UniDim dimensions may serve as an effective interface for relations of different frameworks. The detailed results for the source and target tasks are shown in Tables~\ref{body-total-pdtb-dim} and~\ref{body-rst-transfer-results} in Appendix~\ref{transfer-learning}. 

\begin{table}[h!]\tiny
\centering
\resizebox{0.9\linewidth}{!}{%
\begin{tabular} {|p{25mm}|p{17mm}|p{15mm}|}
\hline  
\tiny Task
&\tiny Acc.
&\tiny Macro-F1 \\
\hline
\tiny target RST (BERT+Dim) & \underline{\textbf{0.81}} & \underline{\textbf{0.57}} \\ \hline
\tiny RST-specific (BERT+Dim) from Table~\ref{body-rst-cls-dim-report} & 0.81 & 0.58 \\ \hline
\tiny src PDTB total (BERT+Dim) & 0.86 & 0.67  \\ \hline
\tiny target RST (BERT only) & 0.58 & 0.33 \\ \hline
\tiny RST-specific (BERT only) from Table~\ref{body-rst-cls-dim-report} & 0.63 & 0.47 \\ \hline
\tiny src PDTB total (BERT only) & 0.71 (vs. DISRPT 2021: 0.74) & 0.61  \\ \hline
\end{tabular} %
}
\vspace{2mm}
\caption{\label{transfer-learning-pdtb2rst} Results of transfer learning from PDTB to RST. }
\end{table}

\subsubsection{Automatic Dimension Prediction}
We show our experimental results of automatic prediction of UniDim dimensions in Table~\ref{dim-pred-results}. As is clear from the table, reasonable performance for this task can be achieved. Note that the baseline results are based on PDTB 2.0 and separate classifiers are trained for each dimension. 

The performance on PDTB is higher than on RST data with the exception of \textit{Temporality} and \textit{Goal}. As PDTB allows multi-sense annotation, instances labeled with temporal relations might be annotated with labels of causal relations, and instances for which a \textit{Purpose} relation can be inferred (captured by the \textit{Goal} dimension), a \textit{Manner} relation is also possible (not involving the \textit{Goal} dimension), which poses a challenge for machine learning systems.

Moreover, combining the two corpora to augment training data does not improve the performance over using PDTB data alone but it is helpful for improving performance on RST data. RST data amount is much smaller and adding more data is beneficial. As relations of the two frameworks may not be completely compatible and combining the two corpora might introduce inconsistent and redundant data, combining the datasets is likely to be more useful in low-resource settings.    

\begin{table}[h!]\tiny
\centering
\resizebox{0.9\linewidth}{!}{%
\begin{tabular} {|p{0.5cm}| >{\columncolor[RGB]{230, 242, 255}} p{0.3 cm}|>{\columncolor[gray]{0.9}}p{0.5 cm}|>{\columncolor[RGB]{230, 242, 255}} p{0.3 cm}| >{\columncolor[gray]{0.9}}p{0.5 cm}|>{\columncolor[RGB]{230, 242, 255}}p{0.3cm}|>{\columncolor[gray]{0.9}}p{0.5cm}|p{0.3cm}|p{0.5cm}|}
\hline  
 &  \multicolumn{2}{c|} {\tiny PDTB}  
 &  \multicolumn{2}{c|} {\tiny RST}
 &  \multicolumn{2}{c|} {\tiny PDTB+RST} 
 &  \multicolumn{2}{c|} {\tiny Baseline}  \\ 
\hline
   & \tiny Acc. & \tiny Macro-F1 & \tiny Acc. & \tiny Macro-F1 & \tiny Acc. & \tiny Macro-F1 &\tiny Acc. & \tiny Macro-F1   \\    \hline
\tiny Pol. & 0.92 & 0.57 & 0.85 & 0.58 &  0.89&  0.56 &  0.82  &  0.50  \\    \hline
\tiny Basic Op. & 0.80 & 0.52 & 0.76 & 0.45 & 0.77 & 0.50 &  0.76  & 0.38 \\    \hline
\tiny SoC. & 0.75 & 0.72 & 0.67 & 0.45 & 0.70 & 0.59  &   0.68 & 0.50 \\   \hline
\tiny Impl. order & 0.76 & 0.50 & 0.75 & 0.38 & 0.75 & 0.48 &  0.78 & 0.41 \\   \hline
\tiny Temp. & 0.79 & 0.59 & 0.86 & 0.30 & 0.82 & 0.43 &  0.73   &  0.48 \\ \hline
\tiny Spec. & 0.87 & 0.65 & 0.80 &  0.72 & 0.83 & 0.66  &  0.85   &  - \\ \hline
\tiny Alter. & 1.00 & 0.95 & 1.00 & 0.50 & 1.00 & 0.95 &  0.99 &  -  \\ \hline
\tiny Cond. & 0.99 & 0.86 & 0.98 & 0.83 & 0.98 & 0.83  &  0.99 &  -   \\ \hline
\tiny Goal & 0.91 & 0.75 & 0.97 & 0.75 & 0.93 & 0.74 & - &  - \\ \hline

\hline
\end{tabular} %
}
\vspace{2mm}
\caption{\label{dim-pred-results} Results of UniDim dimension prediction. Blue columns show classification accuracy and grey columns show macro-averaged F1.}
\end{table}

\section{Conclusion and Future Work}
By incorporating the UniDim dimensions proposed in~\citet{sanders2018unifying} in discourse relation classification tasks, we obtain quantitative results of the effectiveness of these dimensions in capturing discourse relations of different frameworks and bridging discourse relations across frameworks. Ablation studies reveal the influence of these dimensions on different types of discourse relations. Meanwhile, we show that these dimensions can be predicted automatically with a simple model. These dimensions are potentially useful features for discourse relation classification across frameworks. Therefore, in future work, we plan to incorporate automatically predicted dimensions in our models.



\section{Limitations}
Since we need to create the mapping table for PDTB 3.0 by ourselves, it is unavoidable that there may be errors and inconsistencies with existing mapping tables for the other frameworks. 

Meanwhile, in the mapping table provided in~\citet{sanders2018unifying}, to obtain the values of the dimensions, we need all the information of a relation label, for instance, to represent an RST relation label with dimensions, we need the nuclearity label and whether the relation is mono-nuclear or multi-nuclear in addition to the relation label itself, and in the case of a PDTB relation, we need the relation label and the order of the arguments. This is because these dimensions are not incorporated in the annotation process of RST-DT and PDTB, and only a general mapping is possible. We consider the resultant ambiguity and under-specification unavoidable.

\section{Ethics Statement}
This study does not involve special ethical considerations. The potential impact may include providing computational evidence of the validity of cognitive study of discourse relations and attracting attention to cognitive frameworks of discourse, which may spur fine-grained research on the correlation between cognitive dimensions and different discourse relations and how different language models perform from this perspective.

\bibliography{anthology,custom}

\begin{thebibliography}{43}
\expandafter\ifx\csname natexlab\endcsname\relax\def\natexlab#1{#1}\fi

\bibitem[{Afantenos et~al.(2012)Afantenos, Asher, Benamara, Bras, Fabre,
  Ho-dac, Draoulec, Muller, P{\'e}ry-Woodley, Pr{\'e}vot, Rebeyrolles, Tanguy,
  Vergez-Couret, and Vieu}]{afantenos-etal-2012-empirical}
Stergos Afantenos, Nicholas Asher, Farah Benamara, Myriam Bras, C{\'e}cile
  Fabre, Mai Ho-dac, Anne~Le Draoulec, Philippe Muller, Marie-Paule
  P{\'e}ry-Woodley, Laurent Pr{\'e}vot, Josette Rebeyrolles, Ludovic Tanguy,
  Marianne Vergez-Couret, and Laure Vieu. 2012.
\newblock \href
  {http://www.lrec-conf.org/proceedings/lrec2012/pdf/836_Paper.pdf} {An
  empirical resource for discovering cognitive principles of discourse
  organisation: the {ANNODIS} corpus}.
\newblock In \emph{Proceedings of the Eighth International Conference on
  Language Resources and Evaluation ({LREC}'12)}, pages 2727--2734, Istanbul,
  Turkey. European Language Resources Association (ELRA).

\bibitem[{Asher et~al.(2016)Asher, Hunter, Morey, Farah, and
  Afantenos}]{asher-etal-2016-discourse}
Nicholas Asher, Julie Hunter, Mathieu Morey, Benamara Farah, and Stergos
  Afantenos. 2016.
\newblock \href {https://aclanthology.org/L16-1432} {Discourse structure and
  dialogue acts in multiparty dialogue: the {STAC} corpus}.
\newblock In \emph{Proceedings of the Tenth International Conference on
  Language Resources and Evaluation ({LREC}'16)}, pages 2721--2727,
  Portoro{\v{z}}, Slovenia. European Language Resources Association (ELRA).

\bibitem[{Asher and Lascarides(2003)}]{asher2003logics}
Nicholas Asher and Alex Lascarides. 2003.
\newblock \emph{Logics of conversation}.
\newblock Cambridge University Press.

\bibitem[{Atwell et~al.(2021)Atwell, Li, and
  Alikhani}]{atwell-etal-2021-discourse}
Katherine Atwell, Junyi~Jessy Li, and Malihe Alikhani. 2021.
\newblock \href {https://aclanthology.org/2021.sigdial-1.34} {Where are we in
  discourse relation recognition?}
\newblock In \emph{Proceedings of the 22nd Annual Meeting of the Special
  Interest Group on Discourse and Dialogue}, pages 314--325, Singapore and
  Online. Association for Computational Linguistics.

\bibitem[{Benamara and Taboada(2015)}]{benamara-taboada-2015-mapping}
Farah Benamara and Maite Taboada. 2015.
\newblock \href {https://doi.org/10.18653/v1/S15-1016} {Mapping different
  rhetorical relation annotations: A proposal}.
\newblock In \emph{Proceedings of the Fourth Joint Conference on Lexical and
  Computational Semantics}, pages 147--152, Denver, Colorado. Association for
  Computational Linguistics.

\bibitem[{Bourgonje and Zolotarenko(2019)}]{bourgonje-zolotarenko-2019-toward}
Peter Bourgonje and Olha Zolotarenko. 2019.
\newblock \href {https://doi.org/10.18653/v1/W19-2702} {Toward cross-theory
  discourse relation annotation}.
\newblock In \emph{Proceedings of the Workshop on Discourse Relation Parsing
  and Treebanking 2019}, pages 7--11, Minneapolis, MN. Association for
  Computational Linguistics.

\bibitem[{Bunt and Prasad(2016)}]{bunt2016iso}
Harry Bunt and Rashmi Prasad. 2016.
\newblock \href
  {https://let.uvt.nl/general/people/bunt/docs/ISA-12-DR-core-revised.pdf}
  {{ISO DR-Core} {(ISO 24617-8)}: Core concepts for the annotation of discourse
  relations}.
\newblock In \emph{Proceedings 12th Joint ACL-ISO Workshop on Interoperable
  Semantic Annotation (ISA-12)}, pages 45--54.

\bibitem[{Carlson et~al.(2001)Carlson, Marcu, and
  Okurovsky}]{carlson-etal-2001-building}
Lynn Carlson, Daniel Marcu, and Mary~Ellen Okurovsky. 2001.
\newblock \href {https://aclanthology.org/W01-1605} {Building a
  discourse-tagged corpus in the framework of {R}hetorical {S}tructure
  {T}heory}.
\newblock In \emph{Proceedings of the Second {SIG}dial Workshop on Discourse
  and Dialogue}.

\bibitem[{Chiarcos(2014)}]{chiarcos-2014-towards}
Christian Chiarcos. 2014.
\newblock \href
  {http://www.lrec-conf.org/proceedings/lrec2014/pdf/893_Paper.pdf} {Towards
  interoperable discourse annotation. discourse features in the ontologies of
  linguistic annotation}.
\newblock In \emph{Proceedings of the Ninth International Conference on
  Language Resources and Evaluation ({LREC}'14)}, pages 4569--4577, Reykjavik,
  Iceland. European Language Resources Association (ELRA).

\bibitem[{Crible and Degand(2019)}]{crible2019domains}
Ludivine Crible and Liesbeth Degand. 2019.
\newblock Domains and functions: A two-dimensional account of discourse
  markers.
\newblock \emph{Discours. Revue de linguistique, psycholinguistique et
  informatique. A journal of linguistics, psycholinguistics and computational
  linguistics}, (24).

\bibitem[{Demberg et~al.(2019)Demberg, Scholman, and
  Asr}]{demberg2019compatible}
Vera Demberg, Merel~CJ Scholman, and Fatemeh~Torabi Asr. 2019.
\newblock \href
  {https://pdfs.semanticscholar.org/351b/aa55ab47b0e0035b84689c912f1384abdcdb.pdf?_ga=2.176636005.1920837913.1640009376-896286635.1640009376}
  {How compatible are our discourse annotation frameworks? insights from
  mapping rst-dt and pdtb annotations}.
\newblock \emph{Dialogue \& Discourse}, 10(1):87--135.

\bibitem[{Devlin et~al.(2019)Devlin, Chang, Lee, and
  Toutanova}]{devlin-etal-2019-bert}
Jacob Devlin, Ming-Wei Chang, Kenton Lee, and Kristina Toutanova. 2019.
\newblock \href {https://doi.org/10.18653/v1/N19-1423} {{BERT}: Pre-training of
  deep bidirectional transformers for language understanding}.
\newblock In \emph{Proceedings of the 2019 Conference of the North {A}merican
  Chapter of the Association for Computational Linguistics: Human Language
  Technologies, Volume 1 (Long and Short Papers)}, pages 4171--4186,
  Minneapolis, Minnesota. Association for Computational Linguistics.

\bibitem[{Forbes et~al.(2003)Forbes, Miltsakaki, Prasad, Sarkar, Joshi, and
  Webber}]{forbes2003d}
Katherine Forbes, Eleni Miltsakaki, Rashmi Prasad, Anoop Sarkar, Aravind Joshi,
  and Bonnie Webber. 2003.
\newblock \href {https://link.springer.com/article/10.1023/A:1024137719751}
  {{D-LTAG} system: Discourse parsing with a lexicalized tree-adjoining
  grammar}.
\newblock \emph{Journal of Logic, Language and Information}, 12(3):261--279.

\bibitem[{Fu(2022)}]{fu-2022-towards}
Yingxue Fu. 2022.
\newblock \href {https://doi.org/10.18653/v1/2022.acl-srw.12} {Towards
  unification of discourse annotation frameworks}.
\newblock In \emph{Proceedings of the 60th Annual Meeting of the Association
  for Computational Linguistics: Student Research Workshop}, pages 132--142,
  Dublin, Ireland. Association for Computational Linguistics.

\bibitem[{Gessler et~al.(2021)Gessler, Behzad, Liu, Peng, Zhu, and
  Zeldes}]{gessler-etal-2021-discodisco}
Luke Gessler, Shabnam Behzad, Yang~Janet Liu, Siyao Peng, Yilun Zhu, and Amir
  Zeldes. 2021.
\newblock \href {https://doi.org/10.18653/v1/2021.disrpt-1.6}
  {{D}is{C}o{D}is{C}o at the {DISRPT}2021 shared task: A system for discourse
  segmentation, classification, and connective detection}.
\newblock In \emph{Proceedings of the 2nd Shared Task on Discourse Relation
  Parsing and Treebanking (DISRPT 2021)}, pages 51--62, Punta Cana, Dominican
  Republic. Association for Computational Linguistics.

\bibitem[{Ji and Huang(2021)}]{ji-huang-2021-discodvt}
Haozhe Ji and Minlie Huang. 2021.
\newblock \href {https://doi.org/10.18653/v1/2021.emnlp-main.347}
  {{D}isco{DVT}: {G}enerating long text with discourse-aware discrete
  variational transformer}.
\newblock In \emph{Proceedings of the 2021 Conference on Empirical Methods in
  Natural Language Processing}, pages 4208--4224, Online and Punta Cana,
  Dominican Republic. Association for Computational Linguistics.

\bibitem[{Ji and Eisenstein(2014)}]{ji-eisenstein-2014-representation}
Yangfeng Ji and Jacob Eisenstein. 2014.
\newblock \href {https://doi.org/10.3115/v1/P14-1002} {Representation learning
  for text-level discourse parsing}.
\newblock In \emph{Proceedings of the 52nd Annual Meeting of the Association
  for Computational Linguistics (Volume 1: Long Papers)}, pages 13--24,
  Baltimore, Maryland. Association for Computational Linguistics.

\bibitem[{Ji and Eisenstein(2015)}]{ji-eisenstein-2015-one}
Yangfeng Ji and Jacob Eisenstein. 2015.
\newblock \href {https://doi.org/10.1162/tacl_a_00142} {One vector is not
  enough: Entity-augmented distributed semantics for discourse relations}.
\newblock \emph{Transactions of the Association for Computational Linguistics},
  3:329--344.

\bibitem[{Kim et~al.(2020)Kim, Feng, Gunasekara, and
  Lastras}]{kim-etal-2020-implicit}
Najoung Kim, Song Feng, Chulaka Gunasekara, and Luis Lastras. 2020.
\newblock \href {https://doi.org/10.18653/v1/2020.acl-main.480} {Implicit
  discourse relation classification: We need to talk about evaluation}.
\newblock In \emph{Proceedings of the 58th Annual Meeting of the Association
  for Computational Linguistics}, pages 5404--5414, Online. Association for
  Computational Linguistics.

\bibitem[{Ko et~al.(2022)Ko, Dalton, Simmons, Fisher, Durrett, and
  Li}]{ko-etal-2022-discourse}
Wei-Jen Ko, Cutter Dalton, Mark Simmons, Eliza Fisher, Greg Durrett, and
  Junyi~Jessy Li. 2022.
\newblock \href {https://aclanthology.org/2022.emnlp-main.806} {Discourse
  comprehension: A question answering framework to represent sentence
  connections}.
\newblock In \emph{Proceedings of the 2022 Conference on Empirical Methods in
  Natural Language Processing}, pages 11752--11764, Abu Dhabi, United Arab
  Emirates. Association for Computational Linguistics.

\bibitem[{Loshchilov and Hutter(2018)}]{loshchilov2018decoupled}
Ilya Loshchilov and Frank Hutter. 2018.
\newblock Decoupled weight decay regularization.
\newblock In \emph{International Conference on Learning Representations}.

\bibitem[{Maat and Sanders(2000)}]{maat2000domains}
Henk~Pander Maat and Ted Sanders. 2000.
\newblock \href
  {https://www.degruyter.com/document/doi/10.1515/9783110219043.1.57/html}
  {Domains of use or subjectivity? the distribution of three dutch causal
  connectives explained}.
\newblock \emph{Topics in English Linguistics}, 33:57--82.

\bibitem[{Mann and Thompson(1988)}]{mann1988rhetorical}
William~C Mann and Sandra~A Thompson. 1988.
\newblock \href
  {https://www.degruyter.com/document/doi/10.1515/text.1.1988.8.3.243/html}
  {Rhetorical structure theory: Toward a functional theory of text
  organization}.
\newblock \emph{Text}, 8(3):243--281.

\bibitem[{Marcu(2000)}]{marcu2000theory}
Daniel Marcu. 2000.
\newblock \emph{The theory and practice of discourse parsing and
  summarization}.
\newblock MIT press.

\bibitem[{Paszke et~al.(2019)Paszke, Gross, Massa, Lerer, Bradbury, Chanan,
  Killeen, Lin, Gimelshein, Antiga et~al.}]{paszke2019pytorch}
Adam Paszke, Sam Gross, Francisco Massa, Adam Lerer, James Bradbury, Gregory
  Chanan, Trevor Killeen, Zeming Lin, Natalia Gimelshein, Luca Antiga, et~al.
  2019.
\newblock Pytorch: An imperative style, high-performance deep learning library.
\newblock \emph{Advances in Neural Information Processing Systems}, 32.

\bibitem[{Pedregosa et~al.(2011)Pedregosa, Varoquaux, Gramfort, Michel,
  Thirion, Grisel, Blondel, Prettenhofer, Weiss, Dubourg, Vanderplas, Passos,
  Cournapeau, Brucher, Perrot, and Duchesnay}]{scikit-learn}
F.~Pedregosa, G.~Varoquaux, A.~Gramfort, V.~Michel, B.~Thirion, O.~Grisel,
  M.~Blondel, P.~Prettenhofer, R.~Weiss, V.~Dubourg, J.~Vanderplas, A.~Passos,
  D.~Cournapeau, M.~Brucher, M.~Perrot, and E.~Duchesnay. 2011.
\newblock Scikit-learn: Machine learning in {P}ython.
\newblock \emph{Journal of Machine Learning Research}, 12:2825--2830.

\bibitem[{Prasad et~al.(2008)Prasad, Dinesh, Lee, Miltsakaki, Robaldo, Joshi,
  and Webber}]{prasad-etal-2008-penn}
Rashmi Prasad, Nikhil Dinesh, Alan Lee, Eleni Miltsakaki, Livio Robaldo,
  Aravind Joshi, and Bonnie Webber. 2008.
\newblock \href
  {http://www.lrec-conf.org/proceedings/lrec2008/pdf/754_paper.pdf} {The {P}enn
  {D}iscourse {T}ree{B}ank 2.0.}
\newblock In \emph{Proceedings of the Sixth International Conference on
  Language Resources and Evaluation ({LREC}'08)}, Marrakech, Morocco. European
  Language Resources Association (ELRA).

\bibitem[{Prasad et~al.(2019)Prasad, Webber, Lee, and Joshi}]{ldcexample}
Rashmi Prasad, Bonnie Webber, Alan Lee, and Aravind Joshi. 2019.
\newblock \href
  {https://catalog.ldc.upenn.edu/desc/addenda/LDC2019T05_examples.html} {Penn
  discourse treebank version 3.0 ldc2019t05}.

\bibitem[{Pyatkin et~al.(2020)Pyatkin, Klein, Tsarfaty, and
  Dagan}]{pyatkin-etal-2020-qadiscourse}
Valentina Pyatkin, Ayal Klein, Reut Tsarfaty, and Ido Dagan. 2020.
\newblock \href {https://doi.org/10.18653/v1/2020.emnlp-main.224}
  {{QAD}iscourse - {D}iscourse {R}elations as {QA} {P}airs: {R}epresentation,
  {C}rowdsourcing and {B}aselines}.
\newblock In \emph{Proceedings of the 2020 Conference on Empirical Methods in
  Natural Language Processing (EMNLP)}, pages 2804--2819, Online. Association
  for Computational Linguistics.

\bibitem[{Rehbein et~al.(2016)Rehbein, Scholman, and
  Demberg}]{rehbein-etal-2016-annotating}
Ines Rehbein, Merel Scholman, and Vera Demberg. 2016.
\newblock \href {https://aclanthology.org/L16-1165} {Annotating discourse
  relations in spoken language: A comparison of the {PDTB} and {CCR}
  frameworks}.
\newblock In \emph{Proceedings of the Tenth International Conference on
  Language Resources and Evaluation ({LREC}'16)}, pages 1039--1046,
  Portoro{\v{z}}, Slovenia. European Language Resources Association (ELRA).

\bibitem[{Roze et~al.(2019)Roze, Braud, and Muller}]{roze-etal-2019-aspects}
Charlotte Roze, Chlo{\'e} Braud, and Philippe Muller. 2019.
\newblock \href {https://doi.org/10.18653/v1/W19-5950} {Which aspects of
  discourse relations are hard to learn? primitive decomposition for discourse
  relation classification}.
\newblock In \emph{Proceedings of the 20th Annual SIGdial Meeting on Discourse
  and Dialogue}, pages 432--441, Stockholm, Sweden. Association for
  Computational Linguistics.

\bibitem[{Sanders et~al.(2018)Sanders, Demberg, Hoek, Scholman, Asr, Zufferey,
  and Evers-Vermeul}]{sanders2018unifying}
Ted~JM Sanders, Vera Demberg, Jet Hoek, Merel~CJ Scholman, Fatemeh~Torabi Asr,
  Sandrine Zufferey, and Jacqueline Evers-Vermeul. 2018.
\newblock \href
  {https://www.degruyter.com/document/doi/10.1515/cllt-2016-0078/html}
  {Unifying dimensions in coherence relations: How various annotation
  frameworks are related}.
\newblock \emph{Corpus Linguistics and Linguistic Theory}.

\bibitem[{Sanders et~al.(1992)Sanders, Spooren, and
  Noordman}]{sanders1992toward}
Ted~JM Sanders, Wilbert~PM Spooren, and Leo~GM Noordman. 1992.
\newblock \href {https://www.tandfonline.com/doi/abs/10.1080/01638539209544800}
  {Toward a taxonomy of coherence relations}.
\newblock \emph{Discourse processes}, 15(1):1--35.

\bibitem[{Sanders et~al.(1993)Sanders, Spooren, and
  Noordman}]{sanders1993coherence}
Ted~JM Sanders, Wilbert~PM Spooren, and Leo~GM Noordman. 1993.
\newblock \href
  {https://www.degruyter.com/document/doi/10.1515/cogl.1993.4.2.93/html}
  {Coherence relations in a cognitive theory of discourse representation}.

\bibitem[{Scheffler and Stede(2016)}]{scheffler2016mapping}
Tatjana Scheffler and Manfred Stede. 2016.
\newblock \href
  {https://www.linguistics.rub.de/konvens16/pub/31_konvensproc.pdf} {Mapping
  {PDTB}-style connective annotation to {RST}-style discourse annotation}.
\newblock In \emph{Proceedings of the 13th Conference on Natural Language
  Processing}, pages 242--247.

\bibitem[{Shi and Demberg(2019)}]{shi-demberg-2019-next}
Wei Shi and Vera Demberg. 2019.
\newblock \href {https://doi.org/10.18653/v1/D19-1586} {Next sentence
  prediction helps implicit discourse relation classification within and across
  domains}.
\newblock In \emph{Proceedings of the 2019 Conference on Empirical Methods in
  Natural Language Processing and the 9th International Joint Conference on
  Natural Language Processing (EMNLP-IJCNLP)}, pages 5790--5796, Hong Kong,
  China. Association for Computational Linguistics.

\bibitem[{Sim~Smith(2017)}]{sim-smith-2017-integrating}
Karin Sim~Smith. 2017.
\newblock \href {https://doi.org/10.18653/v1/W17-4814} {On integrating
  discourse in machine translation}.
\newblock In \emph{Proceedings of the Third Workshop on Discourse in Machine
  Translation}, pages 110--121, Copenhagen, Denmark. Association for
  Computational Linguistics.

\bibitem[{Wang et~al.(2019)Wang, Dai, P{\'o}czos, and
  Carbonell}]{wang2019characterizing}
Zirui Wang, Zihang Dai, Barnab{\'a}s P{\'o}czos, and Jaime Carbonell. 2019.
\newblock \href
  {https://openaccess.thecvf.com/content_CVPR_2019/html/Wang_Characterizing_and_Avoiding_Negative_Transfer_CVPR_2019_paper.html}
  {Characterizing and avoiding negative transfer}.
\newblock In \emph{Proceedings of the IEEE/CVF conference on computer vision
  and pattern recognition}, pages 11293--11302.

\bibitem[{Wolf and Gibson(2004)}]{wolf-gibson-2004-representing}
Florian Wolf and Edward Gibson. 2004.
\newblock \href {https://aclanthology.org/C04-1020} {Representing discourse
  coherence: A corpus-based analysis}.
\newblock In \emph{{COLING} 2004: Proceedings of the 20th International
  Conference on Computational Linguistics}, pages 134--140, Geneva,
  Switzerland. COLING.

\bibitem[{Wolf et~al.(2020)Wolf, Debut, Sanh, Chaumond, Delangue, Moi, Cistac,
  Rault, Louf, Funtowicz, Davison, Shleifer, von Platen, Ma, Jernite, Plu, Xu,
  Le~Scao, Gugger, Drame, Lhoest, and Rush}]{wolf-etal-2020-transformers}
Thomas Wolf, Lysandre Debut, Victor Sanh, Julien Chaumond, Clement Delangue,
  Anthony Moi, Pierric Cistac, Tim Rault, Remi Louf, Morgan Funtowicz, Joe
  Davison, Sam Shleifer, Patrick von Platen, Clara Ma, Yacine Jernite, Julien
  Plu, Canwen Xu, Teven Le~Scao, Sylvain Gugger, Mariama Drame, Quentin Lhoest,
  and Alexander Rush. 2020.
\newblock \href {https://doi.org/10.18653/v1/2020.emnlp-demos.6} {Transformers:
  State-of-the-art natural language processing}.
\newblock In \emph{Proceedings of the 2020 Conference on Empirical Methods in
  Natural Language Processing: System Demonstrations}, pages 38--45, Online.
  Association for Computational Linguistics.

\bibitem[{Yang et~al.(2019)Yang, Dai, Yang, Carbonell, Salakhutdinov, and
  Le}]{yang2019xlnet}
Zhilin Yang, Zihang Dai, Yiming Yang, Jaime Carbonell, Russ~R Salakhutdinov,
  and Quoc~V Le. 2019.
\newblock Xlnet: Generalized autoregressive pretraining for language
  understanding.
\newblock \emph{Advances in Neural Information Processing Systems}, 32.

\bibitem[{Zeldes et~al.(2021)Zeldes, Liu, Iruskieta, Muller, Braud, and
  Badene}]{zeldes-etal-2021-disrpt}
Amir Zeldes, Yang~Janet Liu, Mikel Iruskieta, Philippe Muller, Chlo{\'e} Braud,
  and Sonia Badene. 2021.
\newblock \href {https://doi.org/10.18653/v1/2021.disrpt-1.1} {The {DISRPT}
  2021 shared task on elementary discourse unit segmentation, connective
  detection, and relation classification}.
\newblock In \emph{Proceedings of the 2nd Shared Task on Discourse Relation
  Parsing and Treebanking (DISRPT 2021)}, pages 1--12, Punta Cana, Dominican
  Republic. Association for Computational Linguistics.

\bibitem[{Zhao and Webber(2021)}]{zhao-webber-2021-revisiting}
Zheng Zhao and Bonnie Webber. 2021.
\newblock \href {https://doi.org/10.18653/v1/2021.codi-main.10} {Revisiting
  shallow discourse parsing in the {PDTB}-3: Handling intra-sentential
  implicits}.
\newblock In \emph{Proceedings of the 2nd Workshop on Computational Approaches
  to Discourse}, pages 107--121, Punta Cana, Dominican Republic and Online.
  Association for Computational Linguistics.

\end{thebibliography}
\bibliographystyle{acl_natbib}

\appendix

\clearpage
\onecolumn
\section{RST to UniDim Dimension Mapping Table} \label{rst-to-unidim-mapping-table}

Table~\ref{rst-ccr-mapping-table} shows the mapping of RST-DT relation labels to UniDim dimensions. 
\begin{tiny}
\begin{longtable}{|l|l|l|l|l|l|l|l|l|l|}
\hline  \textbf{Class}
&\tiny  \textbf{End label} 
&\tiny \textbf{Nuc.}
&\tiny \textbf{N-S}
&\tiny \textbf{Pol.}
&\tiny \textbf{Basic Op.}
&\tiny \textbf{Impl. order}
&\tiny \textbf{SoC}
&\tiny \textbf{Temp.}
&\tiny \textbf{Add. features}
\\  \hline
 Background & Background & Mono & N-S & pos/neg & add & N.A. & obj & anti/N.A. & \\ \hline
 & Background & Mono & S-N & pos/neg & add & N.A. & obj & chron/N.A. & \\ \hline
 & Circumstance & Mono &  & pos/neg & add & N.A. & obj & syn/N.A. & \\ \hline
Cause  & Cause & Mono & N-S & pos & cau & bas & obj & chron & \\ \hline
  & Cause & Mono & S-N & pos & cau & non-b & obj & anti & \\ \hline
 & Cause-result & Multi &  & pos & cau & bas/non-b &obj & chron/anti & \\ \hline
 &Result & Mono & N-S & pos & cau & non-b & obj & anti & \\ \hline
 & Result & Mono & S-N & pos & cau & bas & obj & chron & \\ \hline
 & Consequence-n & Mono & N-S & pos & cau & non-b & obj & anti & \\ \hline
  & Consequence-n & Mono & S-N & pos & cau & bas  & obj & chron & \\ \hline
  & Consequence-s & Mono & N-S & pos & cau & bas  & obj & chron & \\ \hline
  & Consequence-s & Mono & S-N & pos & cau & non-b  & obj & anti & \\ \hline
  & Consequence & Multi &  & pos & cau & bas/non-b  & obj & chron/anti & \\ \hline
Comparison  & Comparison & Both &  & pos & add & N.A.  & obj/sub & N.A. & \\ \hline
        & Preference & Mono &  & neg & add & N.A.  & obj/sub & N.A. & \\ \hline
        & Analogy & Both &  & pos & add & N.A.  & sub & N.A. & \\ \hline
       & Proportion & Multi &  & pos & add/cau & any & obj/sub & any & \\ \hline
Conditional & Condition & Mono & N-S & pos/neg & cau & non-b & obj/sub & anti/N.A. & conditional \\ \hline
    & Condition & Mono & S-N & pos/neg & cau & bas & obj/sub & chron/N.A. & conditional \\ \hline
        & Hypothetical & Mono & N-S & pos & cau & non-b & sub & N.A. & conditional \\ \hline
         & Hypothetical & Mono & S-N & pos & cau & bas & sub & N.A. & conditional \\ \hline
        & Contingency & Mono & N-S & pos/neg & cau & non-b & obj & anti & conditional \\ \hline
        & Contingency & Mono & S-N & pos/neg & cau & bas & obj & chron & conditional \\ \hline
         & Otherwise & Mono & N-S & neg & cau & bas & obj/sub & chron/N.A. & conditional \\ \hline
        & Otherwise & Multi &  & neg & cau & bas & obj/sub & chron/N.A. & conditional \\ \hline
Contrast  & Contrast & Multi &  & neg & add & N.A. & obj/sub & any &   \\ \hline
          & Concession & Mono & N-S & neg & cau & non-b & obj/sub & anti/N.A. & \\ \hline
        & Concession & Mono & S-N & neg & cau & bas & obj/sub & chron/N.A. & \\ \hline
        & Antithesis & Mono &  & neg & add/cau & any & obj/sub & any &  \\ \hline
Elaboration & El.-additional & Mono &   & pos & add & N.A. & obj/sub & N.A. &  \\ \hline
       & El.-gen.-spec. & Mono &   & pos & add & N.A. & obj/sub & N.A. & specificity \\ \hline
       & El.-part-whole & Mono &   & pos & add & N.A. & obj & N.A. & specificity \\ \hline
       & El.-process-step & Mono &   & pos & add & N.A. & obj & N.A. & specificity \\ \hline
       & El.-object-attr. & Mono &   & pos & add & N.A. & obj & N.A. & specificity \\ \hline
       & El.-set-member & Mono &   & pos & add & N.A. & obj & N.A. & spec.-ex. \\ \hline
        & Example & Mono &   & pos & add & N.A. & obj & N.A. & spec.-ex. \\ \hline
        & Definition & Mono &   & pos & add & N.A. & obj & N.A. & specificity \\ \hline

Enablement & Purpose & Mono & N-S & pos & cau & bas & obj/sub & chron/N.A. & goal \\ \hline
            & Purpose & Mono & S-N & pos & cau & non-b & obj/sub & anti/N.A. & goal \\ \hline
           & Enablement & Mono & N-S & pos & cau & non-b & obj/sub & anti/N.A. & goal \\ \hline
            & Enablement & Mono & S-N & pos & cau & bas & obj/sub & chron/N.A. & goal \\ \hline

Evaluation & Evaluation & Both &   & pos & add/cau & any & sub & N.A. & specificity\\ \hline
           & Interpretation & Both &   & pos & add/cau & any & sub & N.A. & specificity\\ \hline
           & Conclusion & Mono & N-S  & pos & cau & bas & sub & N.A. & specificity\\ \hline
           & Conclusion & Mono & S-N  & pos & cau & non-b & sub & N.A. & specificity\\ \hline
          & Conclusion & Multi &   & pos & cau & bas/non-b & sub & N.A. & specificity\\ \hline
         & Comment  & Mono &   & pos & add & N.A. & sub & N.A. & specificity\\ \hline
Explanation & Evidence & Mono & N-S  & pos & cau & non-b & sub & anti &  \\ \hline
            & Evidence & Mono & S-N  & pos & cau & bas & sub & chron &  \\ \hline
            & Exp.-argument. & Mono & N-S  & pos & cau & non-b & obj & anti &  \\ \hline
            & Exp.-argument. & Mono & S-N  & pos & cau & bas & obj & chron &  \\ \hline
            & Reason & Mono & N-S  & pos & cau & non-b & obj & anti &  \\ \hline
            & Reason & Mono & S-N  & pos & cau & bas & obj & chron &  \\ \hline
            & Reason & Multi &  & pos & cau & bas/non-b & obj & chron/anti &  \\ \hline
Joint & List & Multi &  & pos & add & N.A. & obj/sub & syn/chron/N.A. & list \\ \hline
     & Disjunction & Multi & & pos/neg & add & N.A. & obj/sub & syn/N.A. & alternative  \\ \hline
Summary & Summary & Mono &  & pos & add & N.A. & obj & N.A. & specificity \\ \hline
        & Restatement & Mono &  & pos & add & N.A. & obj & N.A. & spec.-equiv. \\ \hline
Temporal  & Temp.-before & Mono & N-S & pos & add & N.A. & obj & chron &  \\ \hline
         & Temp.-before & Mono & S-N & pos & add & N.A. & obj & anti &  \\ \hline
        & Temp.-after & Mono & N-S & pos & add & N.A. & obj & anti &  \\ \hline
        & Temp.-after & Mono & S-N & pos & add & N.A. & obj & chron &  \\ \hline
        & Temp.-same-time & Both &  & pos & add & N.A. & obj & syn &  \\ \hline
        & Sequence & Multi &  & pos & add & N.A. & obj & chron &  \\ \hline
        & Inverted-seq. & Multi &  & pos & add & N.A. & obj & anti &  \\ \hline

 Manner-Means & Means & Mono & N-S & pos & cau & non-b & obj & anti &  \\ \hline
              & Means & Mono & S-N & pos & cau & bas & obj & chron &  goal \\ \hline
 Topic-Comment & Problem-sol.-n & Mono & N-S & pos & cau & non-b & obj/sub & anti/N.A. &  goal \\ \hline
               & Problem-sol.-n & Mono & S-N & pos & cau & bas & obj/sub & chron/N.A. &  goal \\ \hline
                & Problem-sol.-s & Mono & N-S & pos & cau & bas & obj/sub & chron/N.A. &  goal \\ \hline
                & Problem-sol.-s & Mono & S-N & pos & cau & non-b & obj/sub & anti/N.A. &  goal \\ \hline
                & Problem-sol. & Multi &  & pos & cau & bas/non-b & obj/sub & achron/anti/N.A. &  goal \\ \hline

\caption{\label{rst-ccr-mapping-table} Mapping of RST relations to UniDim dimensions, taken from~\citet{sanders2018unifying}}
\end{longtable}
\end{tiny}
Table~\ref{rst-ccr-mapping-table} is the mapping table of relation labels of RST-DT to UniDim dimensions.~\textit{\textbf{Nuc.}} means the nuclearity of a relation. ~\textit{\textbf{N-S}} means whether the nuclearity is Nucleus-Satellite (N-S) or Satellite-Nucleus (S-N) or Nucleus-Nucleus (N-N). ~\textit{\textbf{Pol.}},~\textit{\textbf{Basic Op.}},~\textit{\textbf{Impl. order}},~\textit{\textbf{Basic Op.}},~\textit{\textbf{SoC}},~\textit{\textbf{Temp.}}, and~\textit{\textbf{Add. features}} denote polarity, basic operation, source of coherence, temporality and additional features, respectively.

\clearpage
\twocolumn


\clearpage
\onecolumn
\section{Relation Labels of PDTB 3.0 to UniDim Dimension Mapping Table} \label{pdtb-to-unidim-mapping-table}

Table~\ref{pdtb-ccr-mapping-table} shows the mapping of relation labels of PDTB 3.0 to UniDim dimensions. 
\begin{tiny}
\begin{longtable}{|l|l|l|l|l|l|l|l|l|}
\hline  \textbf{Class\_type}
&\tiny  \textbf{End label} 
&\tiny \textbf{A1-A2}
&\tiny \textbf{Pol.}
&\tiny \textbf{Basic Op.}
&\tiny \textbf{Impl. order}
&\tiny \textbf{SoC}
&\tiny \textbf{Temp.}
&\tiny \textbf{Add. features}
\\ 
\hline
\textbf{Temporal}& & & &  & & &   & \\ \hline
Synchronous&  &  &pos & add &N.A. &obj & sync & \\ \hline
Asynchronous & Precedence & A1-A2 & pos & add & N.A. & obj & chron & \\ \hline
             & Precedence & A2-A1 & pos & add & N.A. & obj & anti & \\ \hline
             & Succession & A1-A2 & pos & add & N.A. & obj & anti & \\ \hline
             & Succession & A2-A1 & pos & add & N.A. & obj & chron & \\ \hline

\textbf{Contingency}& & & &  & & &   & \\ \hline

 Cause & Reason & A1-A2 & pos & cau & non-b & obj & anti & \\ \hline
       & Reason & A2-A1 & pos & cau & bas & obj & chron & \\ \hline
       & Result & A1-A2 & pos & cau & bas & obj & chron &  goal \\ \hline
       & Result & A1-A2 & pos & cau & bas & obj & chron &  goal \\ \hline
       & NegResult &  & neg & cau & bas & obj & chron &     \\ \hline

Cause+Belief & Reason+Belief & A1-A2 & pos & cau & non-b & sub & NS &     \\ \hline
             & Reason+Belief & A2-A1 & pos & cau & bas & sub & NS &  \\ \hline
            & Result+Belief & A1-A2 & pos & cau & bas & sub & NS &     \\ \hline
            & Result+Belief & A2-A1 & pos & cau & non-b & sub & NS &     \\ \hline

\makecell{Cause\\+SpeechAct} & Reason+SpeechAct & A1-A2 & pos & cau & non-b & sub & NS &     \\ \hline
                & Reason+SpeechAct & A2-A1 & pos & cau & bas & sub & NS &     \\ \hline
                & Result+SpeechAct & A1-A2 & pos & cau & bas & sub & NS &     \\ \hline
                & Result+SpeechAct & A2-A1 & pos & cau & non-b & sub & NS &     \\ \hline
Purpose & arg1-as-goal & A1-A2 & pos & cau & non-b & obj/sub & NS & goal   \\ \hline
        & arg1-as-goal & A2-A1 & pos & cau & bas & obj/sub & NS & goal   \\ \hline
        & arg2-as-goal & A1-A2 & pos & cau & bas & sub & NS & goal   \\ \hline

Condition & arg1-as-cond & A1-A2 & pos & cau & bas & obj/sub & NS & conditional  \\ \hline
          & arg1-as-cond & A2-A1 & pos & cau & non-b & obj/sub & NS & conditional  \\ \hline
          & arg2-as-cond & A1-A2 & pos & cau & non-b & obj/sub & NS & conditional  \\ \hline
          & arg2-as-cond & A2-A1 & pos & cau & bas & obj/sub & NS & conditional  \\ \hline

\makecell{Condition\\+SpeechAct} &  &  & pos & cau & bas & sub & NS & conditional  \\ \hline
 
\makecell{Negative\\-Condition} & arg1-as-negcond & A1-A2 & neg & cau & bas & sub & NS & conditional  \\ \hline
                    & arg1-as-negcond & A2-A1 & neg & cau & non-b & sub & NS & conditional  \\ \hline
                    & arg2-as-negcond & A1-A2 & neg & cau & non-b & sub & NS & conditional  \\ \hline
                    & arg2-as-negcond & A2-A1 & neg & cau & bas & sub & NS & conditional  \\ \hline

\makecell{Negative-\\Condition+\\SpeechAct} &  &  & neg & cau & bas & sub & NS & conditional  \\ \hline

\textbf{Comparison}& & & &  & & &   & \\ \hline

Concession & arg1-as-denier & A1-A2 & neg & cau & non-b & obj/sub & NS &  \\ \hline
           & arg1-as-denier & A2-A1 & neg & cau & bas & obj/sub & NS &  \\ \hline
           & arg2-as-denier & A1-A2 & neg & cau & bas & obj/sub & NS &  \\ \hline
           & arg2-as-denier & A2-A1 & neg & cau & non-b & obj/sub & NS &  \\ \hline

\makecell{Concession\\+SpeechAct} &   &   & neg & cau & bas & sub & NS &  \\ \hline
Contrast &   &   & neg & add & NA & obj & NS &  \\ \hline
Similarity &   &   & pos & add & NA & obj & NS &  \\ \hline

\textbf{Expansion}& & & &  & & &   & \\ \hline
Conjunction &   &   & pos & add & NA & obj/sub & NS &  \\ \hline
Disjunction &   &   & neg & add & NA & obj/sub & NS & alternative \\ \hline

Equivalence &   &   & pos & add & NA & obj/sub & NS &  \\ \hline

Exception & arg1-as-excpt & & neg & add & NA & obj/sub & NS &   \\ \hline
          & arg2-as-excpt & & neg & add & NA & obj/sub & NS &  \\ \hline

Instantiation & arg1-as-instance &  & pos & add & NA & obj/sub & NS & specificity \\ \hline
              & arg2-as-instance & & pos & add & NA & obj/sub & NS & specificity \\ \hline
Level-of-detail & arg1-as-detail & & pos & add & NA & obj/sub & NS & specificity \\ \hline
                & arg2-as-detail & & pos & add & NA & obj/sub & NS & specificity \\ \hline
Manner & arg1-as-manner & A1-A2 & pos & add & NA & obj/sub & NS & specificity \\ \hline
       & arg2-as-manner &  & pos & add & NA & obj/sub & NS & specificity \\ \hline

Substitution & arg1-as-subst & A1-A2 & neg & cau & bas & obj/sub & NS & \\ \hline
             & arg1-as-subst & A2-A1 & neg & cau & non-b & obj/sub & NS & \\ \hline
             & arg2-as-subst & A1-A2 & neg & cau & non-b & obj/sub & NS & \\ \hline
             & arg2-as-subst & A2-A1 & neg & cau & bas & obj/sub & NS & \\ \hline

\caption{\label{pdtb-ccr-mapping-table} Mapping of relations labels of PDTB 3.0 to UniDim dimensions.}
\end{longtable}
\end{tiny}
Table~\ref{pdtb-ccr-mapping-table} is the mapping table of relation labels of PDTB 3.0 to UniDim dimensions. A1-A2 means Argument 1 precedes Argument 2 and A2-A1 means Argument 2 precedes Argument 1 in the original text. The abbreviations are interpreted in the same way as in Table~\ref{rst-ccr-mapping-table}.

\clearpage

\section{Distribution of UniDim dimensions in RST-DT and PDTB 3.0}\label{unidim-distributions}

\begin{figure*}[h!]
\noindent\begin{minipage}{\linewidth}
\centering
\includegraphics[
  width=0.8\textwidth,height=0.7\textheight, angle=-90, scale=0.8]{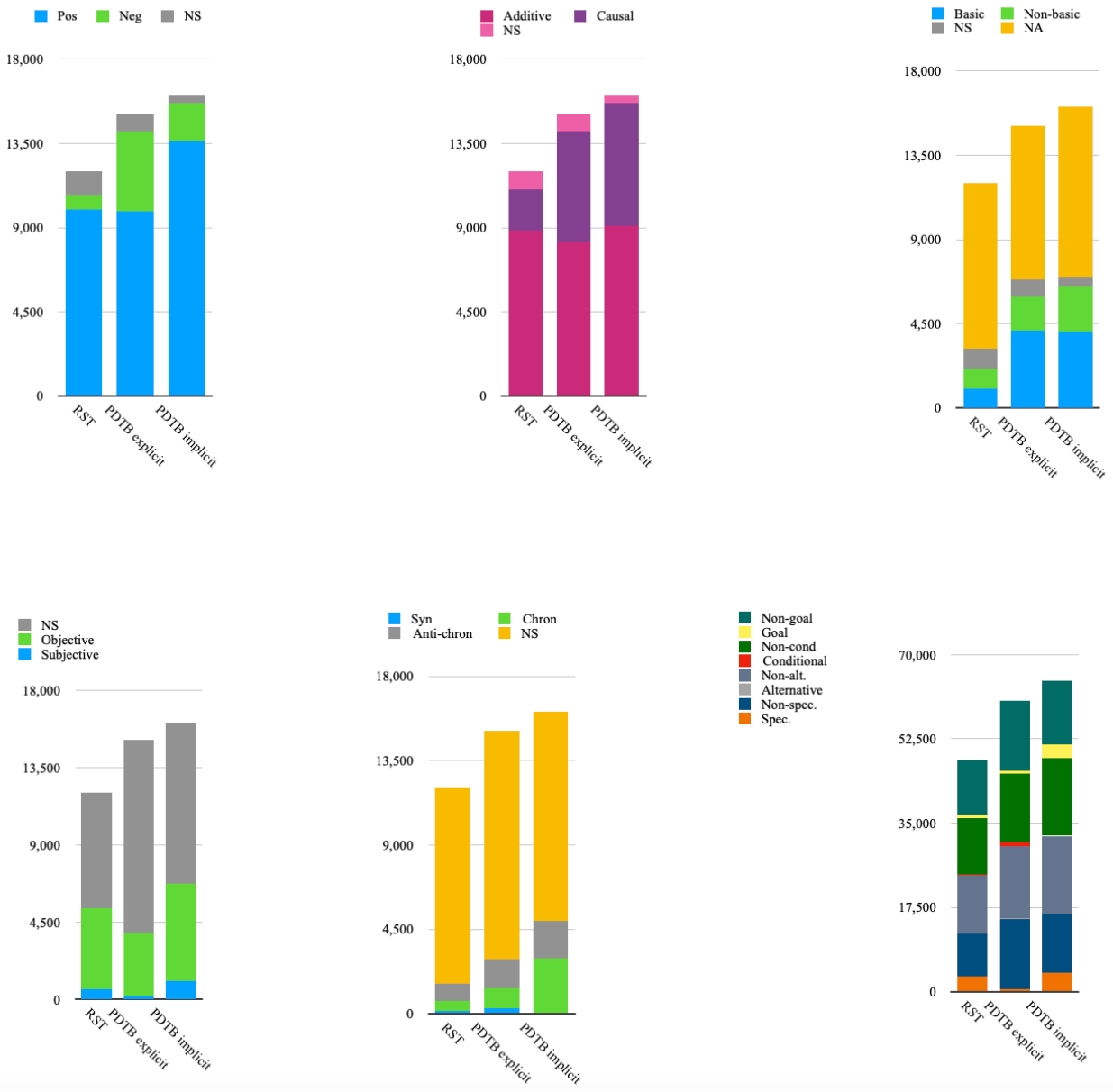}
 \vspace{1\baselineskip}
  \caption{Distribution of the polarity, basic operation, and implication order dimensions (upper row, from left to right, respectively), and source of coherence, temporality and additional dimensions (lower row, from left to right, respectively) in the training sets of RST-DT and PDTB 3.0. We divide PDTB 3.0 based on explicit and implicit relation types. }
  \label{all-dim-dist}
\end{minipage}
\end{figure*}

Figure~\ref{all-dim-dist} shows distribution of the polarity, basic operation, implication order, source of coherence, temporality and additional dimensions used in this paper.  

\clearpage
\twocolumn

\section{Hyper-parameters} \label{hyperparameter}
For discourse relation classification described in section~\ref{section:3.1}, the model is configured with a dropout rate of 0.2. The size of the output of the first MLP is set to 256 and the size of the second MLP output is 128. The model is trained with the AdamW optimizer~\citep{loshchilov2018decoupled}, with a learning rate of $5e-5$. The batch size is set to 4 and the maximum norm of gradient clipping is set to 1. We use get\_linear\_schedule\_with\_warmup from the Transformers library as the learning rate scheduler. The maximum training epoch number is set to 10. The same setting is used in training the model for UniDim dimension prediction, the only exception being the learning rate, which is set to $1e-5$ to obtain good performance for this task. 

For the cross-framework discourse relation classification task, the learning rate for transfer learning is $1e-5$ and as only parameters of the classifier layer are learnable, the maximum training epoch number is set to 50. The other hyper-parameters are the same as above.

We choose the best-performing model based on the performance at the validation set. The PyTorch library~\citep{paszke2019pytorch} is used for implementation. The models are trained on an RTX2060 Super GPU.

The model for PDTB relation classification has 109,753,388 parameters and the training process took 6:25:23 (h:mm:ss) GPU hours for PDTB total relation classification, 2:56:58 GPU hours for PDTB explicit relation classification and 3:13:13 GPU hours for PDTB implicit relation classification. The model for RST relation classification has 109,494,544 parameters and the training process took 2:28:44 GPU hours. The number of parameters in the model for transfer learning is 2,064 and the training process took 4:38:43 GPU hours. 

\clearpage
\onecolumn

\section{Distribution of Relations in Training Data} \label{rel-train-stats}
\begin{figure*}[h!]
\centering
  \includegraphics [scale = 0.35]{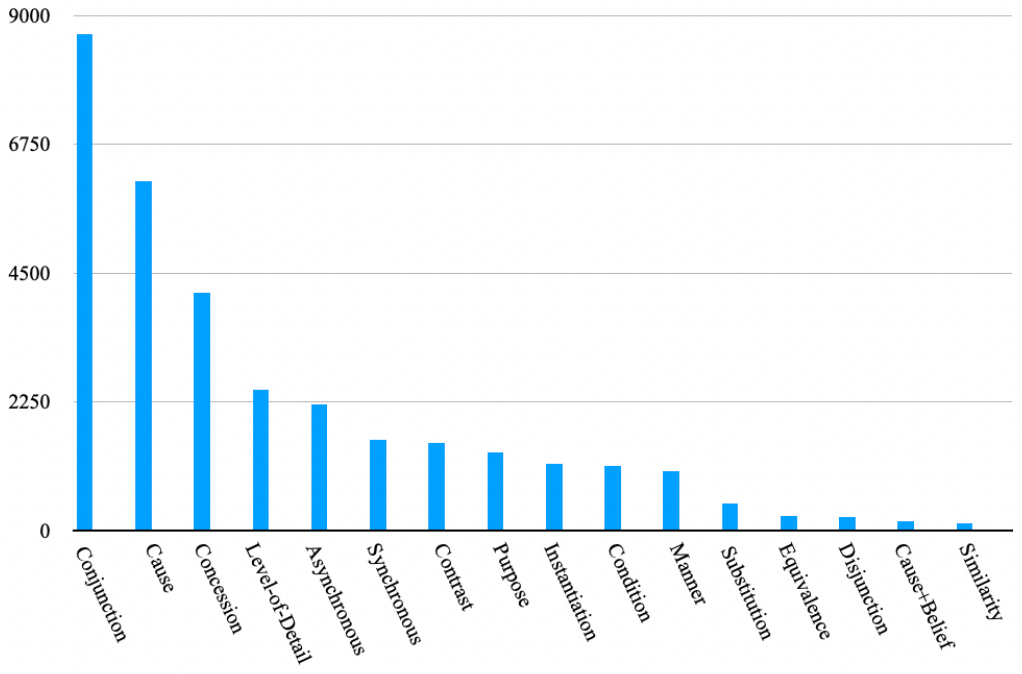}
  \vspace{-1cm}
  \caption{Distribution of PDTB relations in the experiment on PDTB where data of explicit and implicit relations are combined.}
  \label{pdtb-total-train-label-stats}
\end{figure*}

\begin{figure*}[h!]
  \centering
  \includegraphics[scale = 0.35]{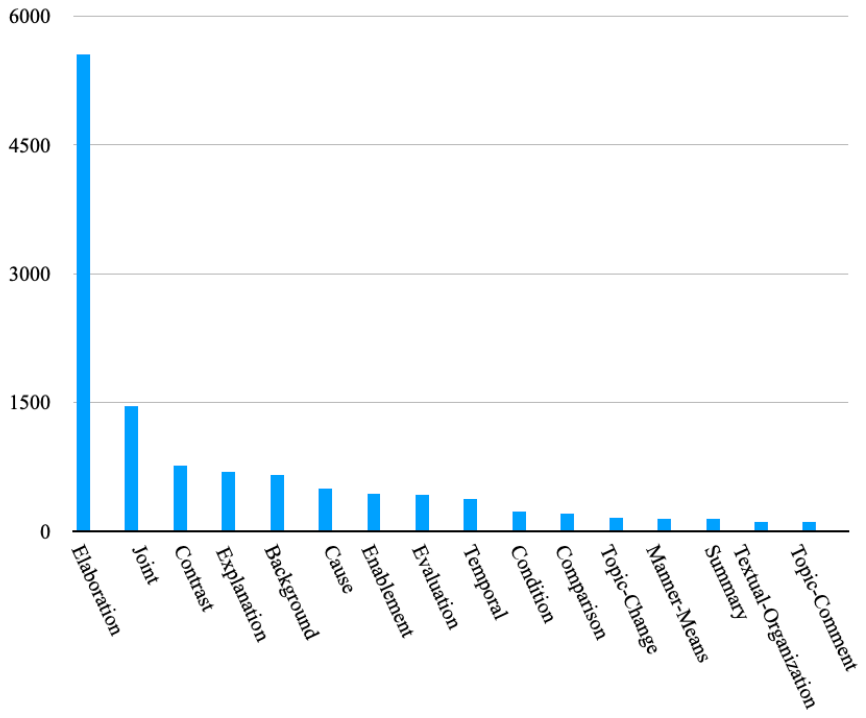}
  \vspace{-1cm}
  \caption{Distribution of RST relations in the training set.}
  \label{rst-train-stats}
\end{figure*}

\begin{figure*}[h!]
  \centering
  \includegraphics[scale = 0.35]{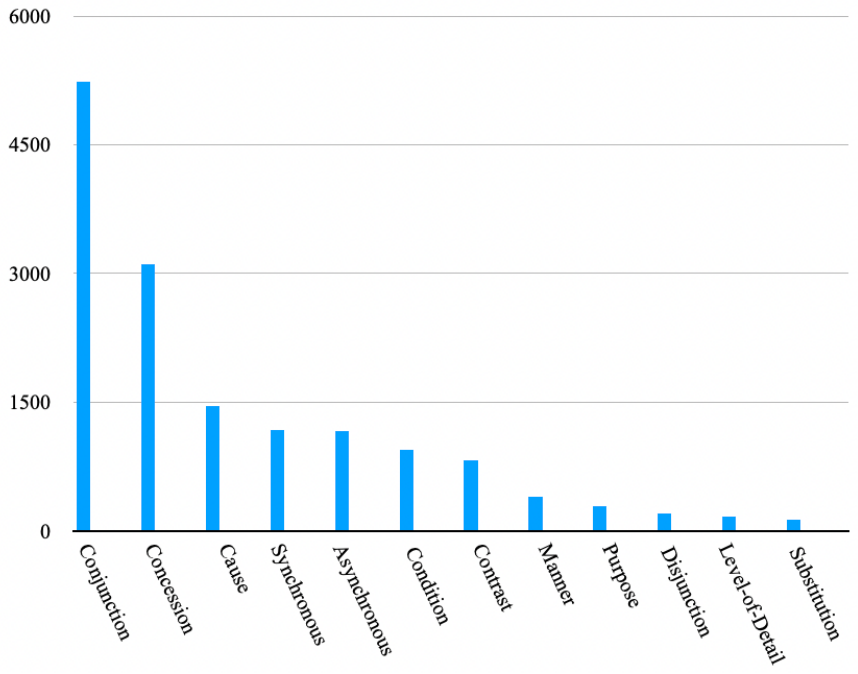}
  \vspace{-1cm}
  \caption{Distribution of PDTB explicit relations in the training set.}
  \label{pdtb-explicit-train-stats}
\end{figure*}

\begin{figure*}[h!]
  \centering
  \includegraphics[scale = 0.35]{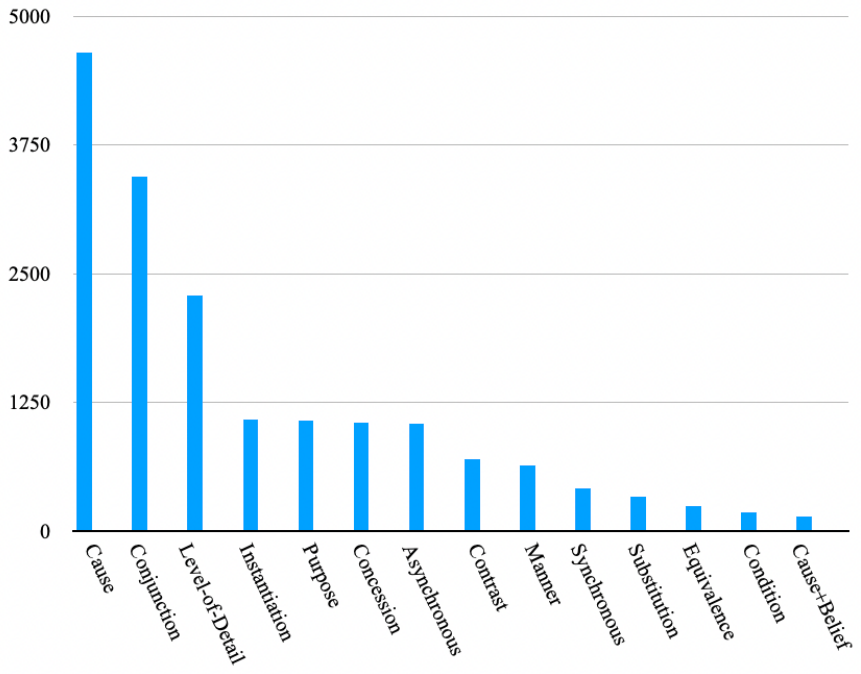}
  \vspace{-1cm}
  \caption{Distribution of PDTB implicit relations in the training set.}
  \label{pdtb-implicit-train-stats}
\end{figure*}

Figures~\ref{pdtb-total-train-label-stats},~\ref{rst-train-stats},~\ref{pdtb-explicit-train-stats} and~\ref{pdtb-implicit-train-stats} shows the distribution of relations in the training sets used in the experiments, sorted in descending order.  

\clearpage
\twocolumn

\section{PDTB Total Data Relation Classification} \label{pdtb-total-cls-result}

Table~\ref{total-pdtb-dim} shows the classification report on PDTB 3.0 (combining explicit and implicit relations) with BERT embeddings and UniDim dimensions as input features. 

\begin{table}[H]\tiny
\centering
\begin{tabular}{|l|l|l|l|l|}
\hline  
&\tiny Precision 
&\tiny Recall 
&\tiny F1
&\tiny Support\\ 
\hline
 Asynchronous & 1.00 & 1.00 & 1.00 & 232\\ \hline
Cause & 1.00 & 1.00 & 1.00 & 538 \\ \hline
Cause+Belief & 1.00 & 1.00 & 1.00 & 13 \\ \hline
Concession & 0.99 & 0.96 & 0.98 & 371 \\ \hline
Condition & 1.00 & 1.00 & 1.00 & 79 \\ \hline
Conjunction & 0.97 & 1.00 & 0.98 & 745 \\ \hline
Contrast & 1.00 & 1.00 & 1.00 & 102 \\ \hline
Disjunction & 1.00 & 1.00 & 1.00 & 20 \\ \hline
Equivalence & 0.00 & 0.00 & 0.00 & 25 \\ \hline
Instantiation & 0.00 & 0.00 & 0.00 & 117 \\ \hline
Level-of-detail & 0.00 & 0.00 & 0.00 & 202 \\ \hline
Manner & 0.07 & 0.96 & 0.14 & 26 \\ \hline
Purpose & 1.00 & 0.96 & 0.98 & 118 \\ \hline
Similarity & 0.00 & 0.00 & 0.00 & 12 \\ \hline
Substitution & 0.68 & 0.91 & 0.78 & 35 \\ \hline
Synchronous & 0.90 & 1.00 & 0.95 & 170 \\ \hline
\textbf{Accuracy}& \multicolumn{4}{|c|} {\textbf{0.86}} \\ \hline
\textbf{Macro-F1} & \textbf{0.66} & \textbf{0.74} & \textbf{0.67} & 2805 \\ 
\hline
\end{tabular}
\vspace{2mm}
\caption{\label{total-pdtb-dim} PDTB relation classification with BERT embeddings and UniDim dimensions as features.}
\end{table}

Table~\ref{total-pdtb-bert} shows the classification report on PDTB 3.0 (combining explicit and implicit relations) with BERT embeddings as input. 

\begin{table}[H]\tiny
\centering
\begin{tabular}{|l|l|l|l|l|}
\hline  
&\tiny Precision 
&\tiny Recall 
&\tiny F1
&\tiny Support\\ 
\hline
 Asynchronous & 0.79 & 0.65 & 0.71 & 232\\ \hline
Cause & 0.71 & 0.62 & 0.66 & 538 \\ \hline
Cause+Belief & 0.00 & 0.00 & 0.00 & 13 \\ \hline
Concession & 0.78 & 0.83 & 0.80 & 371 \\ \hline
Condition & 0.92 & 0.87 & 0.90 & 79 \\ \hline
Conjunction & 0.71 & 0.85 & 0.77 & 745 \\ \hline
Contrast & 0.48 & 0.40 & 0.44 & 102 \\ \hline
Disjunction & 0.86 & 0.90 & 0.88 & 20 \\ \hline
Equivalence & 0.36 & 0.16 & 0.22 & 25 \\ \hline
Instantiation & 0.70 & 0.57 & 0.63 & 117 \\ \hline
Level-of-detail & 0.48 & 0.53 & 0.50 & 202 \\ \hline
Manner & 0.41 & 0.62 & 0.49 & 26 \\ \hline
Purpose & 0.87 & 0.84 & 0.85 & 118 \\ \hline
Similarity & 0.78 & 0.58 & 0.67 & 12 \\ \hline
Substitution & 0.53 & 0.49 & 0.51 & 35 \\ \hline
Synchronous & 0.74 & 0.64 & 0.68 & 170 \\ \hline
\textbf{Accuracy}& \multicolumn{4}{|c|} {\textbf{0.71}} \\ \hline
\textbf{Macro-F1} & \textbf{0.63} & \textbf{0.60} & \textbf{0.61} & 2805 \\ 
\hline
\end{tabular}
\vspace{2mm}
\caption{\label{total-pdtb-bert} PDTB relation classification with BERT embeddings as features.}
\end{table}

\section{PDTB Explicit Relation Classification}\label{pdtb-explicit-cls-table}

Table~\ref{expl-pdtb-dim} shows the classification report on PDTB 3.0 (explicit relations only) with BERT embeddings and UniDim dimensions as input features. 

\begin{table}[H]\tiny
\centering
\begin{tabular}{|l|l|l|l|l|}
\hline  
&\tiny Precision 
&\tiny Recall 
&\tiny F1
&\tiny Support\\ 
\hline
 Asynchronous & 1.00 & 1.00 & 1.00 & 127\\ \hline
Cause & 1.00 & 1.00 & 1.00 & 115 \\ \hline
Concession & 0.96 & 1.00 & 0.98 & 285 \\ \hline
Condition & 1.00 & 1.00 & 1.00 & 61 \\ \hline
Conjunction & 1.00 & 1.00 & 1.00 & 516 \\ \hline
Contrast & 1.00 & 1.00 & 1.00 & 50 \\ \hline
Disjunction & 1.00 & 1.00 & 1.00 & 18 \\ \hline
Level-of-detail & 0.00 & 0.00 & 0.00 & 20 \\ \hline
Manner & 0.35 & 1.00 & 0.52 & 11 \\ \hline
Purpose & 1.00 & 1.00 & 1.00 & 29 \\ \hline
Substitution & 0.00 & 0.00 & 0.00 & 13 \\ \hline
Synchronous & 1.00 & 1.00 & 1.00 & 126 \\ \hline
\textbf{Accuracy}& \multicolumn{4}{|c|} {\textbf{0.98}} \\ \hline
\textbf{Macro-F1} & \textbf{0.78} & \textbf{0.83} & \textbf{0.79} & 1371 \\ 
\hline
\end{tabular}
\vspace{2mm}
\caption{\label{expl-pdtb-dim} Classification report of PDTB explicit relations with BERT embeddings and UniDim dimensions as features.}
\end{table}

Table~\ref{expl-pdtb-bert-only} shows the classification report on PDTB 3.0 (explicit relations only) with BERT embeddings as input features. 

\begin{table}[H]\tiny
\centering
\begin{tabular}{|l|l|l|l|l|}
\hline  
&\tiny Precision 
&\tiny Recall 
&\tiny F1
&\tiny Support\\ 
\hline
 Asynchronous & 0.97 & 0.87 & 0.92 & 127\\ \hline
Cause & 0.82 & 0.89 & 0.85 & 115 \\ \hline
Concession & 0.89 & 0.95 & 0.92 & 285 \\ \hline
Condition & 0.93 & 0.92 & 0.93 & 61 \\ \hline
Conjunction & 0.97 & 0.96 & 0.96 & 516 \\ \hline
Contrast & 0.52 & 0.48 & 0.50 & 50 \\ \hline
Disjunction & 0.90 & 1.00 & 0.95 & 18 \\ \hline
Level-of-detail & 0.71 & 0.75 & 0.73 & 20 \\ \hline
Manner & 0.42 & 0.91 & 0.57 & 11 \\ \hline
Purpose & 0.62 & 0.45 & 0.52 & 29 \\ \hline
Substitution & 1.00 & 0.92 & 0.96 & 13 \\ \hline
Synchronous & 0.81 & 0.71 & 0.76 & 126 \\ \hline
\textbf{Accuracy}& \multicolumn{4}{|c|} {\textbf{0.89}} \\ \hline
\textbf{Macro-F1} & \textbf{0.80} & \textbf{0.82} & \textbf{0.80} & 1371 \\ 
\hline
\end{tabular}
\vspace{2mm}
\caption{\label{expl-pdtb-bert-only} Classification report of PDTB explicit relations with BERT embeddings as features.}
\end{table}

\section{PDTB Explicit Relation Classification Ablation Studies}\label{pdtb-explicit-cls-ablation-studies}

Table~\ref{expl-pdtb-pol-ab} shows the classification report on PDTB 3.0 (explicit relations only) with BERT embeddings and UniDim dimensions as input features, the polarity dimension being removed. 
\begin{table}[H]\tiny
\centering
\begin{tabular}{|l|l|l|l|l|}
\hline  
&\tiny Precision 
&\tiny Recall 
&\tiny F1
&\tiny Support\\ 
\hline
 Asynchronous & 1.00 & 1.00 & 1.00 & 127\\ \hline
Cause & 1.00 & 1.00 & 1.00 & 115 \\ \hline
Concession & 0.96 & 1.00 & 0.98 & 285 \\ \hline
Condition & 1.00 & 1.00 & 1.00 & 61 \\ \hline
Conjunction & 1.00 & 1.00 & 1.00 & 516 \\ \hline
Contrast & 0.62 & 1.00 & 0.76 & 50 \\ \hline
Disjunction & 1.00 & 1.00 & 1.00 & 18 \\ \hline
Level-of-detail & 0.00 & 0.00 & 0.00 & 20 \\ \hline
Manner & 0.35 & 1.00 & 0.52 & 11 \\ \hline
Purpose & 1.00 & 1.00 & 1.00 & 29 \\ \hline
Substitution & 0.00 & 0.00 & 0.00 & 13 \\ \hline
Synchronous & 1.00 & 0.75 & 0.86 & 126 \\ \hline
\textbf{Accuracy}& \multicolumn{4}{|c|} {\textbf{0.95}} \\ \hline
\textbf{Macro-F1} & \textbf{0.74} & \textbf{0.81} & \textbf{0.76} & 1371 \\ 
\hline
\end{tabular}
\vspace{2mm}
\caption{\label{expl-pdtb-pol-ab} Classification report of PDTB explicit relations, with the polarity dimension removed.}
\end{table}

Table~\ref{expl-pdtb-bop-ab} shows the classification report on PDTB 3.0 (explicit relations only) with BERT embeddings and UniDim dimensions as input features, the basic operation dimension being removed. 
\begin{table}[H]\tiny
\centering
\begin{tabular}{|l|l|l|l|l|}
\hline  
&\tiny Precision 
&\tiny Recall 
&\tiny F1
&\tiny Support\\ 
\hline
 Asynchronous  &     1.00  &    1.00   &   1.00   &    127 \\ \hline
          Cause   &    1.00  &    1.00  &    1.00  &     115 \\ \hline
     Concession   &   0.96   &   1.00  &    0.98  &     285 \\ \hline
      Condition   &    1.00   &   1.00  &     1.00 &       61 \\ \hline
    Conjunction   &    1.00  &    1.00  &    1.00  &     516 \\ \hline
       Contrast   &     1.00  &    1.00  &    1.00  &      50 \\ \hline
    Disjunction   &    1.00  &    1.00  &    1.00  &      18 \\ \hline
Level-of-detail  &     0.00  &     0.00 &     0.00  &      20 \\ \hline
         Manner  &     0.35  &   1.00  &    0.52   &     11 \\ \hline
        Purpose  &     1.00  &    1.00 &     1.00  &      29 \\ \hline
   Substitution   &    0.00  &    0.00 &     0.00  &      13 \\ \hline
    Synchronous  &     1.00  &    1.00  &    1.00  &     126 \\ \hline
    
\textbf{Accuracy}& \multicolumn{4}{|c|} {\textbf{0.98}} \\ \hline
\textbf{Macro-F1} & \textbf{0.78} & \textbf{0.83} & \textbf{0.79} & 1371 \\ 
\hline
\end{tabular}
\vspace{2mm}
\caption{\label{expl-pdtb-bop-ab} Classification report of PDTB explicit relations, with the basic operation dimension removed.}
\end{table}

Table~\ref{expl-pdtb-soc-ab} shows the classification report on PDTB 3.0 (explicit relations only) with BERT embeddings and UniDim dimensions as input features, the source of coherence dimension being removed. 
\begin{table}[H]\tiny
\centering
\begin{tabular}{|l|l|l|l|l|}
\hline  
&\tiny Precision 
&\tiny Recall 
&\tiny F1
&\tiny Support\\ 
\hline
   Asynchronous  &     1.00  &    1.00 &     1.00   &    127 \\ \hline
          Cause  &     1.00   &   1.00  &    1.00    &   115 \\ \hline
     Concession  &    0.96   &   1.00   &   0.98  &      285 \\ \hline
      Condition  &    1.00   &  1.00    &  1.00 &       61 \\ \hline
    Conjunction  &     0.94  &    1.00   &   0.97 &      516 \\ \hline
       Contrast  &     0.74  &    1.00 &    0.85  &      50 \\ \hline
    Disjunction  &     0.00   &   0.00  &    0.00  &      18 \\ \hline
Level-of-detail  &     0.00  &    0.00  &    0.00   &     20 \\ \hline
         Manner  &    0.35  &    1.00   &   0.52   &     11 \\ \hline
        Purpose  &     1.00 &     1.00  &    1.00 &       29 \\ \hline
   Substitution  &     0.00 &     0.00   &   0.00   &     13 \\ \hline
    Synchronous  &      1.00  &    0.75  &    0.86  &     126 \\ \hline

\textbf{Accuracy}& \multicolumn{4}{|c|} {\textbf{0.94}} \\ \hline
\textbf{Macro-F1} & \textbf{0.67} & \textbf{0.73} & \textbf{0.68} & 1371 \\ 
\hline
\end{tabular}
\vspace{2mm}
\caption{\label{expl-pdtb-soc-ab} Classification report of PDTB explicit relations, with the source of coherence dimension removed.}
\end{table}

Table~\ref{expl-pdtb-impl-order-ab} shows the classification report on PDTB 3.0 (explicit relations only) with BERT embeddings and UniDim dimensions as input features, the implication order dimension being removed. 
\begin{table}[H]\tiny
\centering
\begin{tabular}{|l|l|l|l|l|}
\hline  
&\tiny Precision 
&\tiny Recall 
&\tiny F1
&\tiny Support\\ 
\hline
 
  Asynchronous  &     1.00  &    1.00    &  1.00  &     127 \\ \hline
          Cause  &     1.00  &    1.00  &    1.00  &     115 \\ \hline
     Concession  &     0.96  &    1.00   &   0.98   &    285 \\ \hline
      Condition  &     1.00  &    1.00 &     1.00  &      61 \\ \hline
    Conjunction  &     1.00  &    1.00  &    1.00  &     516 \\ \hline
       Contrast  &     1.00  &    1.00  &    1.00  &      50 \\ \hline
    Disjunction  &     1.00  &    1.00  &   1.00  &      18 \\ \hline
Level-of-detail  &     0.00  &    0.00  &    0.00   &     20 \\ \hline
         Manner  &     0.35  &    1.00  &    0.52    &    11 \\ \hline
        Purpose  &    1.00   &   1.00   &   1.00     &   29  \\ \hline
   Substitution   &    0.00  &    0.00   &   0.00    &    13 \\ \hline
    Synchronous   &    1.00   &   1.00   &   1.00    &   126  \\ \hline
\textbf{Accuracy}& \multicolumn{4}{|c|} {\textbf{0.98 }} \\ \hline
\textbf{Macro-F1} & \textbf{0.78} & \textbf{0.83} & \textbf{0.79} & 1371 \\ 
\hline
\end{tabular}
\vspace{2mm}
\caption{\label{expl-pdtb-impl-order-ab} Classification report of PDTB explicit relations, with the implication order dimension removed.}
\end{table}

Table~\ref{expl-pdtb-temp-ab} shows the classification report on PDTB 3.0 (explicit relations only) with BERT embeddings and UniDim dimensions as input features, the temporality dimension being removed. 
\begin{table}[H]\tiny
\centering
\begin{tabular}{|l|l|l|l|l|}
\hline  
&\tiny Precision 
&\tiny Recall 
&\tiny F1
&\tiny Support\\ 
\hline
  Asynchronous  &     0.80  &    1.00  &    0.89  &     127 \\ \hline
          Cause  &     1.00  &    1.00   &   1.00  &     115 \\ \hline
     Concession  &     0.96  &    1.00  &    0.98  &     285 \\ \hline
      Condition  &     1.00  &    1.00  &    1.00  &      61 \\ \hline
    Conjunction  &     1.00   &   1.00  &     1.00  &     516 \\ \hline
       Contrast  &    1.00  &    1.00   &   1.00   &     50 \\ \hline
    Disjunction  &     1.00 &     1.00  &    1.00   &     18 \\ \hline
Level-of-detail  &     0.00  &    0.00   &   0.00    &    20 \\ \hline
         Manner  &     0.35  &    1.00   &   0.52    &    11 \\ \hline
        Purpose  &     1.00  &    1.00   &   1.00   &     29 \\ \hline
   Substitution  &     0.00  &    0.00  &    0.00   &     13 \\ \hline
    Synchronous   &    1.00  &    0.75   &   0.86   &    126 \\ \hline

\textbf{Accuracy}& \multicolumn{4}{|c|} {\textbf{0.95}} \\ \hline
\textbf{Macro-F1} & \textbf{0.76} & \textbf{0.81} & \textbf{0.77} & 1371 \\ 
\hline
\end{tabular}
\vspace{2mm}
\caption{\label{expl-pdtb-temp-ab} Classification report of PDTB explicit relations, with the temporality dimension removed.}
\end{table}

Table~\ref{expl-pdtb-add-feat-ab} shows the classification report on PDTB 3.0 (explicit relations only) with BERT embeddings and UniDim dimensions as input features, the additional dimensions being removed. 
\begin{table}[H]\tiny
\centering
\begin{tabular}{|l|l|l|l|l|}
\hline  
&\tiny Precision 
&\tiny Recall 
&\tiny F1
&\tiny Support\\ 
\hline
   Asynchronous  &      1.00  &    1.00  &    1.00   &    127 \\ \hline
          Cause   &    1.00   &   1.00 &     1.00   &    115 \\ \hline
     Concession   &    0.96  &    1.00  &    0.98  &     285  \\ \hline
      Condition   &   0.88   &   1.00  &    0.94   &     61   \\ \hline
    Conjunction   &    0.94  &     1.00  &    0.97  &     516  \\ \hline
       Contrast  &     1.00  &    1.00  &     1.00  &      50  \\ \hline
    Disjunction  &     1.00  &    1.00  &   1.00    &    18  \\ \hline
Level-of-detail  &     0.00  &    0.00  &    0.00  &      20 \\ \hline
         Manner  &     0.00  &    0.00  &     0.00  &      11  \\ \hline
        Purpose  &      1.00  &    0.72  &     0.84  &      29 \\ \hline
   Substitution  &     0.00   &   0.00   &   0.00   &      13  \\ \hline
    Synchronous  &     1.00   &  1.00  &    1.00  &     126  \\ \hline

\textbf{Accuracy}& \multicolumn{4}{|c|} {\textbf{0.96}} \\ \hline
\textbf{Macro-F1} & \textbf{0.73} & \textbf{0.73} & \textbf{0.73} & 1371 \\ 
\hline
\end{tabular}
\vspace{2mm}
\caption{\label{expl-pdtb-add-feat-ab} Classification report of PDTB explicit relations, with the additional dimensions removed.}
\end{table}

\section{PDTB Implicit Relation Classification} \label{pdtb-impl-cls-table}

Table~\ref{impl-pdtb-dim} shows the classification report on PDTB 3.0 (implicit relations only) with BERT embeddings and UniDim dimensions as input features. 

\begin{table}[H]\tiny
\centering
\begin{tabular}{|l|l|l|l|l|}
\hline  
&\tiny Precision 
&\tiny Recall 
&\tiny F1
&\tiny Support\\ 
\hline
 Asynchronous & 1.00 & 1.00 & 1.00 & 95\\ \hline
Cause & 1.00 & 1.00 & 1.00 & 366\\ \hline
Cause+Belief & 1.00 & 0.42 & 0.59 & 12 \\ \hline
Concession & 1.00 & 0.92 & 0.96 & 84 \\ \hline
Condition & 1.00 & 1.00 & 1.00 & 12 \\ \hline
Conjunction & 0.90 & 1.00 & 0.95 & 221 \\ \hline
Contrast & 0.98 & 1.00 & 0.99 & 50 \\ \hline
Equivalence & 0.00 & 0.00 & 0.00 & 24 \\ \hline
Instantiation & 0.00 & 0.00 & 0.00 & 107 \\ \hline
Level-of-detail & 0.60 & 1.00 & 0.75 & 180 \\ \hline
Manner & 0.00 & 0.00 & 0.00 & 15 \\ \hline
Purpose & 0.92 & 0.94 & 0.93 & 88 \\ \hline
Substitution & 0.75 & 1.00 & 0.86 & 21 \\ \hline
Synchronous & 0.87 & 0.97 & 0.92 & 40 \\ \hline
\textbf{Accuracy}& \multicolumn{4}{|c|} {\textbf{0.87}} \\ \hline
\textbf{Macro-F1} & \textbf{0.72} & \textbf{0.73} & \textbf{0.71} & 1315 \\ 
\hline
\end{tabular}
\vspace{2mm}
\caption{\label{impl-pdtb-dim} Classification report of implicit PDTB relations with BERT embeddings and UniDim dimensions as features.}
\end{table}

Table~\ref{impl-pdtb-bert-only} shows the classification report on PDTB 3.0 (implicit relations only) with only BERT embeddings as input features. 

\begin{table}[H]\tiny
\centering
\begin{tabular}{|l|l|l|l|l|}
\hline  
&\tiny Precision 
&\tiny Recall 
&\tiny F1
&\tiny Support\\ 
\hline
 Asynchronous & 0.62 & 0.61 & 0.62 & 95\\ \hline
Cause & 0.60 & 0.63 & 0.61 & 366\\ \hline
Cause+Belief & 0.00 & 0.00 & 0.00 & 12 \\ \hline
Concession & 0.44 & 0.40 & 0.42 & 84 \\ \hline
Condition & 0.71 & 0.42 & 0.53 & 12 \\ \hline
Conjunction & 0.49 & 0.61 & 0.54 & 221 \\ \hline
Contrast & 0.45 & 0.42 & 0.43 & 50 \\ \hline
Equivalence & 0.12 & 0.04 & 0.06 & 24 \\ \hline
Instantiation & 0.77 & 0.54 & 0.64 & 107 \\ \hline
Level-of-detail & 0.45 & 0.48 & 0.46 & 180 \\ \hline
Manner & 0.38 & 0.60 & 0.46 & 15 \\ \hline
Purpose & 0.92 & 0.98 & 0.95 & 88 \\ \hline
Substitution & 0.43 & 0.48 & 0.45 & 21 \\ \hline
Synchronous & 0.27 & 0.10 & 0.15 & 40 \\ \hline
\textbf{Accuracy}& \multicolumn{4}{|c|} {\textbf{0.56}} \\ \hline
\textbf{Macro-F1} & \textbf{0.48} & \textbf{0.45} & \textbf{0.45} & 1315 \\ 
\hline
\end{tabular}
\vspace{2mm}
\caption{\label{impl-pdtb-bert-only} Classification report of PDTB implicit relations with only BERT embeddings as features.}
\end{table}

\section{PDTB Implicit Relation Classification Ablation Studies}\label{pdtb-impl-cls-ablation-studies}

Table~\ref{impl-pdtb-pol-ab} shows the classification report on PDTB 3.0 (implicit relations only) with BERT embeddings and UniDim dimensions as input features, the polarity dimension being removed. 

\begin{table}[H]\tiny
\centering
\begin{tabular}{|l|l|l|l|l|}
\hline  
&\tiny Precision 
&\tiny Recall 
&\tiny F1
&\tiny Support\\ 
\hline
 Asynchronous & 1.00 & 1.00 & 1.00 & 95\\ \hline
Cause & 1.00 & 1.00 & 1.00 & 366\\ \hline
Cause+Belief & 1.00 & 0.42 & 0.59 & 12 \\ \hline
Concession & 0.96 & 0.92 & 0.94 & 84 \\ \hline
Condition & 1.00 & 0.75 & 0.86 & 12 \\ \hline
Conjunction & 0.90 & 1.00 & 0.95 & 221 \\ \hline
Contrast & 0.98 & 1.00 & 0.99 & 50 \\ \hline
Equivalence & 0.00 & 0.00 & 0.00 & 24 \\ \hline
Instantiation & 0.00 & 0.00 & 0.00 & 107 \\ \hline
Level-of-detail & 0.60 & 1.00 & 0.75 & 180 \\ \hline
Manner & 0.00 & 0.00 & 0.00 & 15 \\ \hline
Purpose & 0.92 & 0.94 & 0.93 & 88 \\ \hline
Substitution & 0.75 & 1.00 & 0.86 & 21 \\ \hline
Synchronous & 0.87 & 0.97 & 0.92 & 40 \\ \hline
\textbf{Accuracy}& \multicolumn{4}{|c|} {\textbf{0.87}} \\ \hline
\textbf{Macro-F1} & \textbf{0.71} & \textbf{0.71} & \textbf{0.70} & 1315 \\ 
\hline
\end{tabular}
\vspace{2mm}
\caption{\label{impl-pdtb-pol-ab} Classification report of PDTB implicit relations, with the polarity dimension removed.}
\end{table}

Table~\ref{impl-pdtb-bop-ab} shows the classification report on PDTB 3.0 (implicit relations only) with BERT embeddings and UniDim dimensions as input features, the basic operation dimension being removed. 

\begin{table}[H]\tiny
\centering
\begin{tabular}{|l|l|l|l|l|}
\hline  
&\tiny Precision 
&\tiny Recall 
&\tiny F1
&\tiny Support\\ 
\hline
 Asynchronous & 1.00 & 1.00 & 1.00 & 95\\ \hline
Cause & 1.00 & 1.00 & 1.00 & 366\\ \hline
Cause+Belief & 1.00 & 0.42 & 0.59 & 12 \\ \hline
Concession & 1.00 & 0.92 & 0.96 & 84 \\ \hline
Condition & 1.00 & 1.00 & 1.00 & 12 \\ \hline
Conjunction & 0.90 & 1.00 & 0.95 & 221 \\ \hline
Contrast & 1.00 & 1.00 & 1.00 & 50 \\ \hline
Equivalence & 0.00 & 0.00 & 0.00 & 24 \\ \hline
Instantiation & 0.00 & 0.00 & 0.00 & 107 \\ \hline
Level-of-detail & 0.60 & 1.00 & 0.75 & 180 \\ \hline
Manner & 0.00 & 0.00 & 0.00 & 15 \\ \hline
Purpose & 0.92 & 0.94 & 0.93 & 88 \\ \hline
Substitution & 0.75 & 1.00 & 0.86 & 21 \\ \hline
Synchronous & 0.87 & 0.97 & 0.92 & 40 \\ \hline
\textbf{Accuracy}& \multicolumn{4}{|c|} {\textbf{0.87}} \\ \hline
\textbf{Macro-F1} & \textbf{0.72} & \textbf{0.73} & \textbf{0.71} & 1315 \\ 
\hline
\end{tabular}
\vspace{2mm}
\caption{\label{impl-pdtb-bop-ab} Classification report of PDTB implicit relations, with the basic operation dimension removed.}
\end{table}

Table~\ref{impl-pdtb-soc-ab} shows the classification report on PDTB 3.0 (implicit relations only) with BERT embeddings and UniDim dimensions as input features, the source of coherence dimension being removed. 

\begin{table}[H]\tiny
\centering
\begin{tabular}{|l|l|l|l|l|}
\hline  
&\tiny Precision 
&\tiny Recall 
&\tiny F1
&\tiny Support\\ 
\hline
 Asynchronous & 1.00 & 1.00 & 1.00 & 95\\ \hline
Cause & 1.00 & 1.00 & 1.00 & 366\\ \hline
Cause+Belief & 1.00 & 0.42 & 0.59 & 12 \\ \hline
Concession & 1.00 & 0.92 & 0.96 & 84 \\ \hline
Condition & 1.00 & 1.00 & 1.00 & 12 \\ \hline
Conjunction & 0.90 & 1.00 & 0.95 & 221 \\ \hline
Contrast & 1.00 & 1.00 & 1.00 & 50 \\ \hline
Equivalence & 0.00 & 0.00 & 0.00 & 24 \\ \hline
Instantiation & 0.00 & 0.00 & 0.00 & 107 \\ \hline
Level-of-detail & 0.60 & 1.00 & 0.75 & 180 \\ \hline
Manner & 0.00 & 0.00 & 0.00 & 15 \\ \hline
Purpose & 0.92 & 0.94 & 0.93 & 88 \\ \hline
Substitution & 0.75 & 1.00 & 0.86 & 21 \\ \hline
Synchronous & 0.87 & 0.97 & 0.92 & 40 \\ \hline
\textbf{Accuracy}& \multicolumn{4}{|c|} {\textbf{0.87}} \\ \hline
\textbf{Macro-F1} & \textbf{0.72} & \textbf{0.73} & \textbf{0.71} & 1315 \\ 
\hline
\end{tabular}
\vspace{2mm}
\caption{\label{impl-pdtb-soc-ab} Classification report of PDTB implicit relations, with the source of coherence dimension removed. The result is the same as Table~\ref{impl-pdtb-bop-ab}, where the basic operation dimension is removed.}
\end{table}

Table~\ref{impl-pdtb-impl-order-ab} shows the classification report on PDTB 3.0 (implicit relations only) with BERT embeddings and UniDim dimensions as input features, the implication order dimension being removed. 

\begin{table}[H]\tiny
\centering
\begin{tabular}{|l|l|l|l|l|}
\hline  
&\tiny Precision 
&\tiny Recall 
&\tiny F1
&\tiny Support\\ 
\hline
 Asynchronous & 1.00 & 1.00 & 1.00 & 95\\ \hline
Cause & 1.00 & 1.00 & 1.00 & 366\\ \hline
Cause+Belief & 0.00 & 0.00 & 0.00 & 12 \\ \hline
Concession & 0.80 & 1.00 & 0.89 & 84 \\ \hline
Condition & 1.00 & 1.00 & 1.00 & 12 \\ \hline
Conjunction & 0.90 & 1.00 & 0.95 & 221 \\ \hline
Contrast & 0.98 & 1.00 & 0.99 & 50 \\ \hline
Equivalence & 0.00 & 0.00 & 0.00 & 24 \\ \hline
Instantiation & 0.00 & 0.00 & 0.00 & 107 \\ \hline
Level-of-detail & 0.60 & 1.00 & 0.75 & 180 \\ \hline
Manner & 0.00 & 0.00 & 0.00 & 15 \\ \hline
Purpose & 0.87 & 0.94 & 0.91 & 88 \\ \hline
Substitution & 0.00 & 0.00 & 0.00 & 21 \\ \hline
Synchronous & 0.87 & 0.97 & 0.92 & 40 \\ \hline
\textbf{Accuracy}& \multicolumn{4}{|c|} {\textbf{0.86}} \\ \hline
\textbf{Macro-F1} & \textbf{0.57} & \textbf{0.64} & \textbf{0.60} & 1315 \\ 
\hline
\end{tabular}
\vspace{2mm}
\caption{\label{impl-pdtb-impl-order-ab} Classification report of PDTB implicit relations, with the implication order dimension removed.}
\end{table}

Table~\ref{impl-pdtb-temp-ab} shows the classification report on PDTB 3.0 (implicit relations only) with BERT embeddings and UniDim dimensions as input features, the temporality dimension being removed. 

\begin{table}[H]\tiny
\centering
\begin{tabular}{|l|l|l|l|l|}
\hline  
&\tiny Precision 
&\tiny Recall 
&\tiny F1
&\tiny Support\\ 
\hline
  Asynchronous  &  0.99  &  1.00   &  0.99    &    95\\ \hline
          Cause &  1.00  &  1.00   &   1.00    &   366\\ \hline
   Cause+Belief & 1.00  &  0.42   &    0.59    &    12\\ \hline
     Concession & 1.00  & 0.92   &   0.96     &   84\\ \hline
      Condition &      1.00  &    1.00  &    1.00   &     12 \\ \hline
    Conjunction  &     0.90  &    1.00  &    0.95  &     221  \\ \hline
       Contrast  &    1.00   &   1.00   &   1.00   &     50 \\ \hline
    Equivalence  &    0.00   &   0.00   &    0.00   &     24 \\ \hline
  Instantiation  &    0.00  &    0.00    &  0.00   &    107\\ \hline
Level-of-detail  &    0.60   &   1.00    &  0.75    &   180  \\ \hline
         Manner  &     0.00   &   0.00    &  0.00     &   15 \\ \hline
        Purpose   &    0.92  &    0.94   &   0.93  &      88 \\ \hline
   Substitution   &    0.75  &    1.00   &   0.86   &     21 \\ \hline
    Synchronous   &    0.87   &   0.97  &    0.92    &    40 \\ \hline

\textbf{Accuracy}& \multicolumn{4}{|c|} {\textbf{0.87}} \\ \hline
\textbf{Macro-F1} & \textbf{0.72} & \textbf{0.73} & \textbf{0.71} & 1315 \\ 
\hline
\end{tabular}
\vspace{2mm}
\caption{\label{impl-pdtb-temp-ab} Classification report of PDTB implicit relations, with the temporality dimension removed.}
\end{table}

Table~\ref{impl-pdtb-add-f-ab} shows the classification report on PDTB 3.0 (implicit relations only) with BERT embeddings and UniDim dimensions as input features, the additional dimensions being removed. 

\begin{table}[H]\tiny
\centering
\begin{tabular}{|l|l|l|l|l|}
\hline  
&\tiny Precision 
&\tiny Recall 
&\tiny F1
&\tiny Support\\ 
\hline
    Asynchronous   &    0.99   &   1.00   &   0.99    &     95 \\ \hline
          Cause    &    1.00   &   1.00   &   1.00    &   366  \\ \hline
   Cause+Belief    &   1.00   &   0.42    &  0.59     &   12   \\ \hline
     Concession    &   0.96   &  0.92   &   0.94     &   84  \\ \hline
      Condition    &    1.00   &   0.75   &   0.86   &     12  \\ \hline
    Conjunction    &   0.40   &   1.00   &   0.58    &   221  \\ \hline
       Contrast    &    1.00   &   1.00  &    1.00   &      50 \\ \hline
    Equivalence    &   0.00  &    0.00  &    0.00  &    24  \\ \hline
  Instantiation    &   0.00  &    0.00  &   0.00  &     107 \\ \hline
Level-of-detail    &   0.00  &   0.00  &    0.00  &     180 \\ \hline
         Manner    &   0.00  &    0.00  &    0.00  &      15 \\ \hline
        Purpose    &    0.92  &    0.94  &    0.93   &     88 \\ \hline
   Substitution    &   0.75   &   1.00   &   0.86    &    21 \\ \hline
    Synchronous    &   0.87   &   0.97   &   0.92   &      40 \\ \hline

\textbf{Accuracy}& \multicolumn{4}{|c|} {\textbf{0.73}} \\ \hline
\textbf{Macro-F1} & \textbf{0.64} & \textbf{0.64} & \textbf{0.62} & 1315 \\ 
\hline
\end{tabular}
\vspace{2mm}
\caption{\label{impl-pdtb-add-f-ab} Classification report of PDTB implicit relations, with the additional dimensions removed.}
\end{table}

\section{RST Relation Classification}\label{rst-rel-cls-result}

Table~\ref{rst-dim-cls-report} shows RST relation classification report with BERT embeddings and UniDim dimensions as input features. 

\begin{table}[h!]\tiny
\centering
\begin{tabular}{|l|l|l|l|l|}
\hline  
&\tiny Precision 
&\tiny Recall 
&\tiny F1
&\tiny Support\\ 
\hline
 Background & 1.00 & 1.00 & 1.00 & 111\\ \hline
Cause & 0.92 & 0.70 & 0.79 & 82 \\ \hline
Comparison & 0.00 & 0.00 & 0.00 & 29 \\ \hline
Condition & 1.00 & 1.00 & 1.00 & 48 \\ \hline
Contrast & 0.99 & 1.00 & 0.99 & 146 \\ \hline
Elaboration & 0.75 & 1.00 & 0.86 & 796 \\ \hline
Enablement & 0.92 & 1.00 & 0.96 & 46 \\ \hline
Evaluation & 0.99 & 1.00 & 0.99 & 80 \\ \hline
Explanation & 0.72 & 0.97 & 0.83 & 110 \\ \hline
Joint & 1.00 & 0.03 & 0.06 & 212 \\ \hline
Manner-Means & 0.00 & 0.00 & 0.00 & 27 \\ \hline
Summary & 0.00 & 0.00 & 0.00 & 32 \\ \hline
Temporal & 1.00 & 1.00 & 1.00 & 73 \\ \hline
Textual-Organization & 0.00 & 0.00 & 0.00 & 9 \\ \hline
Topic-Change & 0.28 & 1.00 & 0.44 & 13 \\ \hline
Topic-Comment & 0.71 & 0.21 & 0.32 & 24 \\ \hline
\textbf{Accuracy}& \multicolumn{4}{|c|} {\textbf{0.81}} \\ \hline
\textbf{Macro-F1} & \textbf{0.64} & \textbf{0.62} & \textbf{0.58} & 1838 \\ 
\hline
\end{tabular}
\vspace{2mm}
\caption{\label{rst-dim-cls-report} RST relation classification report with BERT embeddings and UniDim dimensions as features.}
\end{table}

Table~\ref{rst-bert-cls-report} shows RST relation classification report with BERT embeddings as input features. 

\begin{table}[H]\tiny
\centering
\begin{tabular}{|l|l|l|l|l|}
\hline  
&\tiny Precision 
&\tiny Recall 
&\tiny F1
&\tiny Support\\ 
\hline
 Background & 0.47 & 0.35 & 0.40 & 111\\ \hline
Cause & 0.50 & 0.17 & 0.25 & 82 \\ \hline
Comparison & 0.61 & 0.38 & 0.47 & 29 \\ \hline
Condition & 0.79 & 0.71 & 0.75 & 48 \\ \hline
Contrast & 0.75 & 0.68 & 0.72 & 146 \\ \hline
Elaboration & 0.65 & 0.88 & 0.75 & 796 \\ \hline
Enablement & 0.61 & 0.85 & 0.71 & 46 \\ \hline
Evaluation & 0.29 & 0.14 & 0.19 & 80 \\ \hline
Explanation & 0.46 & 0.27 & 0.34 & 110 \\ \hline
Joint & 0.67 & 0.62 & 0.64 & 212 \\ \hline
Manner-Means & 0.68 & 0.48 & 0.57 & 27 \\ \hline
Summary & 0.88 & 0.47 & 0.61 & 32 \\ \hline
Temporal & 0.74 & 0.27 & 0.40 & 73 \\ \hline
Textual-Organization & 0.44 & 0.44 & 0.44 & 9 \\ \hline
Topic-Change & 0.28 & 0.38 & 0.32 & 13 \\ \hline
Topic-Comment & 0.00 & 0.00 & 0.00 & 24 \\ \hline
\textbf{Accuracy}& \multicolumn{4}{|c|} {\textbf{0.63}} \\ \hline
\textbf{Macro-F1} & \textbf{0.55} & \textbf{0.44} & \textbf{0.47} & 1838 \\ 
\hline
\end{tabular}
\vspace{2mm}
\caption{\label{rst-bert-cls-report} 
RST relation classification report using pre-trained BERT model.}
\end{table}

Table~\ref{rst-dim-transfer} shows RST relation classification report using transfer learning from the PDTB relation classification model (combining PDTB explicit and implicit relation data during training) with BERT embeddings and UnDim dimensions as input features. 

\begin{table}[H]\tiny
\centering
\begin{tabular}{|l|l|l|l|l|}
\hline  
&\tiny Precision 
&\tiny Recall 
&\tiny F1
&\tiny Support\\ 
\hline
 Background & 1.00 & 1.00 & 1.00 & 111\\ \hline
Cause & 0.90 & 0.70 & 0.79 & 82 \\ \hline
Comparison & 0.00 & 0.00 & 0.00 & 29 \\ \hline
Condition & 1.00 & 0.98 & 0.99 & 48 \\ \hline
Contrast & 0.99 & 1.00 & 0.99 & 146 \\ \hline
Elaboration & 0.75 & 1.00 & 0.86 & 796 \\ \hline
Enablement & 0.92 & 1.00 & 0.96 & 46 \\ \hline
Evaluation & 1.00 & 1.00 & 1.00 & 80 \\ \hline
Explanation & 0.72 & 0.97 & 0.83 & 110 \\ \hline
Joint & 0.00 & 0.00 & 0.00 & 212 \\ \hline
Manner-Means & 0.00 & 0.00 & 0.00 & 27 \\ \hline
Summary & 0.00 & 0.00 & 0.00 & 32 \\ \hline
Temporal & 1.00 & 1.00 & 1.00 & 73 \\ \hline
Textual-Organization & 0.00 & 0.00 & 0.00 & 9 \\ \hline
Topic-Change & 0.28 & 1.00 & 0.44 & 13 \\ \hline
Topic-Comment & 0.71 & 0.21 & 0.32 & 24 \\ \hline
\textbf{Accuracy}& \multicolumn{4}{|c|} {\textbf{0.81}} \\ \hline
\textbf{Macro-F1} & \textbf{0.58} & \textbf{0.62} & \textbf{0.57} & 1838 \\ 
\hline
\end{tabular}
\vspace{2mm}
\caption{\label{rst-dim-transfer} 
Transfer learning for RST relation classification with the PDTB relation classification model with BERT embeddings and UniDim dimensions as input features.}
\end{table}

Table~\ref{rst-bert-transfer} shows RST relation classification report using transfer learning from the pre-trained BERT model fine-tuned on PDTB relation classification task (combining PDTB explicit and implicit relation data). 

\begin{table}[H]\tiny
\centering
\begin{tabular}{|l|l|l|l|l|}
\hline  
&\tiny Precision 
&\tiny Recall 
&\tiny F1
&\tiny Support\\ 
\hline
 Background & 0.51 & 0.27 & 0.35 & 111\\ \hline
Cause & 0.17 & 0.07 & 0.10 & 82 \\ \hline
Comparison & 0.42 & 0.38 & 0.40 & 29 \\ \hline
Condition & 0.80 & 0.67 & 0.73 & 48 \\ \hline
Contrast & 0.75 & 0.73 & 0.74 & 146 \\ \hline
Elaboration & 0.60 & 0.82 & 0.69 & 796 \\ \hline
Enablement & 0.48 & 0.78 & 0.60 & 46 \\ \hline
Evaluation & 0.00 & 0.00 & 0.00 & 80 \\ \hline
Explanation & 0.40 & 0.15 & 0.22 & 110 \\ \hline
Joint & 0.57 & 0.66 & 0.61 & 212 \\ \hline
Manner-Means & 0.43 & 0.33 & 0.38 & 27 \\ \hline
Summary & 0.00 & 0.00 & 0.00 & 32 \\ \hline
Temporal & 0.53 & 0.36 & 0.43 & 73 \\ \hline
Textual-Organization & 0.00 & 0.00 & 0.00 & 9 \\ \hline
Topic-Change & 0.00 & 0.00 & 0.00 & 13 \\ \hline
Topic-Comment & 0.00 & 0.00 & 0.00 & 24 \\ \hline
\textbf{Accuracy}& \multicolumn{4}{|c|} {\textbf{0.58}} \\ \hline
\textbf{Macro-F1} & \textbf{0.35} & \textbf{0.33} & \textbf{0.33} & 1838 \\ 
\hline
\end{tabular}
\vspace{2mm}
\caption{\label{rst-bert-transfer} 
Transfer learning for RST relation classification using BERT embeddings as input. }
\end{table}

\section{RST Relation Classification Ablation Studies}\label{rst-rel-cls-ablation-studies}

Table~\ref{rst-pol-ab} shows the classification report on RST-DT with BERT embeddings and UniDim dimensions as input features, the polarity dimension being removed. 

\begin{table}[H]\tiny
\centering
\begin{tabular}{|l|l|l|l|l|}
\hline  
&\tiny Precision 
&\tiny Recall 
&\tiny F1
&\tiny Support\\ 
\hline
   Background    &   1.00  &    1.00  &    1.00   &    111 \\ \hline
               Cause   &    0.90   &   0.70    &  0.79    &    82 \\ \hline
          Comparison   &     0.00  &    0.00   &   0.00    &    29 \\ \hline
           Condition  &     1.00  &    0.94    &  0.97    &    48  \\ \hline
            Contrast  &     0.61  &    0.56  &    0.58    &   146\\ \hline
         Elaboration  &    0.68   &   1.00   &   0.81  &     796 \\ \hline
          Enablement  &     0.92  &     1.00  &     0.96   &     46 \\ \hline
          Evaluation  &     1.00  &    0.57   &   0.73   &     80 \\ \hline
         Explanation  &     0.71  &   0.97   &   0.82  &      110 \\ \hline
               Joint   &    0.00  &    0.00   &   0.00   &    212 \\ \hline
        Manner-Means   &    0.00  &    0.00   &   0.00   &     27 \\ \hline
             Summary   &    0.00  &    0.00   &   0.00    &    32 \\ \hline
            Temporal   &   1.00  &    1.00   &   1.00    &    73 \\ \hline
Textual-organization   &     0.00  &     0.00   &   0.00  &   9 \\ \hline
        Topic-Change  &     0.00   &   0.00  &    0.00    &    13 \\ \hline
       Topic-Comment  &     0.00   &   0.00  &    0.00    &    24 \\ \hline
\textbf{Accuracy}& \multicolumn{4}{|c|} {\textbf{0.74}} \\ \hline
\textbf{Macro-F1} & \textbf{0.49} & \textbf{0.48} & \textbf{0.48} & 1838 \\ 
\hline
\end{tabular}
\vspace{2mm}
\caption{\label{rst-pol-ab} Classification report for RST, with the polarity dimension removed.}
\end{table}

Table~\ref{rst-bop-ab} shows the classification report on RST-DT with BERT embeddings and UniDim dimensions as input features, the basic operation dimension being removed. 

\begin{table}[H]\tiny
\centering
\begin{tabular}{|l|l|l|l|l|}
\hline  
&\tiny Precision 
&\tiny Recall 
&\tiny F1
&\tiny Support\\ 
\hline
    Background     &  0.95   &   1.00  &    0.97   &    111 \\ \hline
               Cause  &     0.90   &   0.70  &     0.79   &     82 \\ \hline
          Comparison  &     0.00   &   0.00  &   0.00    &    29 \\ \hline
           Condition  &     1.00   &   0.98  &    0.99   &     48 \\ \hline
            Contrast  &     0.99  &    1.00  &    0.99   &    146 \\ \hline
         Elaboration  &     0.73  &    1.00   &   0.84  &     796 \\ \hline
          Enablement  &     0.92 &     1.00  &   0.96  &     46 \\ \hline
          Evaluation  &      0.87  &    0.57  &    0.69 &       80 \\ \hline
         Explanation  &     0.72  &    0.97 &     0.83  &      110 \\ \hline
               Joint  &     0.00  &   0.00  &     0.00  &    212 \\ \hline
        Manner-Means  &     0.00  &    0.00 &     0.00  &      27 \\ \hline
             Summary  &     0.00  &   0.00  &     0.00  &      32 \\ \hline
            Temporal  &     1.00  &    1.00 &     1.00  &      73  \\ \hline
Textual-Organization  &    0.00  &    0.00  &     0.00  &       9  \\ \hline
        Topic-Change  &     0.28  &    1.00  &     0.44 &       13 \\ \hline
       Topic-Comment  &      0.00  &     0.00  &    0.00  &      24 \\ \hline

\textbf{Accuracy}& \multicolumn{4}{|c|} {\textbf{0.78}} \\ \hline
\textbf{Macro-F1} & \textbf{0.52} & \textbf{0.58} & \textbf{0.53} & 1838 \\ 
\hline
\end{tabular}
\vspace{2mm}
\caption{\label{rst-bop-ab} Classification report for RST, with the basic operation dimension removed.}
\end{table}

Table~\ref{rst-soc-ab} shows the classification report on RST-DT with BERT embeddings and UniDim dimensions as input features, the source of coherence dimension being removed. 

\begin{table}[H]\tiny
\centering
\begin{tabular}{|l|l|l|l|l|}
\hline  
&\tiny Precision 
&\tiny Recall 
&\tiny F1
&\tiny Support\\ 
\hline
    Background   &    0.95  &    1.00  &    0.97   &    111 \\  \hline
               Cause   &     0.84  &    0.70  &    0.76   &     82 \\ \hline
          Comparison   &    0.00  &    0.00   &   0.00   &      29 \\ \hline
           Condition   &    1.00  &   0.98  &    0.99   &     48 \\ \hline
            Contrast  &     0.99  &    1.00 &     0.99 &      146 \\ \hline
         Elaboration   &    0.73  &    1.00  &    0.84  &     796 \\ \hline
          Enablement   &    0.92  &   1.00  &    0.96 &       46 \\ \hline
          Evaluation   &    0.96  &    0.57 &     0.72 &       80 \\ \hline
         Explanation   &    0.72  &    0.97  &    0.83  &    110 \\ \hline
               Joint   &    0.00  &     0.00  &   0.00  &     212 \\ \hline
        Manner-Means  &     0.00 &     0.00 &     0.00  &      27 \\ \hline
             Summary  &      0.00 &    0.00 &     0.00  &      32 \\ \hline
            Temporal  &     1.00  &     1.00 &     1.00  &       73 \\ \hline
Textual-Organization  &     0.00 &    0.00  &    0.00 &         9 \\ \hline
        Topic-Change  &      0.28 &     1.00 &     0.44  &       13 \\ \hline
       Topic-Comment  &     0.00 &     0.00 &     0.00   &     24 \\ \hline

\textbf{Accuracy}& \multicolumn{4}{|c|} {\textbf{0.78}} \\ \hline
\textbf{Macro-F1} & \textbf{0.52} & \textbf{0.58} & \textbf{0.53} & 1838 \\ 
\hline
\end{tabular}
\vspace{2mm}
\caption{\label{rst-soc-ab} Classification report for RST, with the source of coherence dimension removed.}
\end{table}

Table~\ref{rst-impl-order-ab} shows the classification report on RST-DT with BERT embeddings and UniDim dimensions as input features, the implication order dimension being removed. 

\begin{table}[H]\tiny
\centering
\begin{tabular}{|l|l|l|l|l|}
\hline  
&\tiny Precision 
&\tiny Recall 
&\tiny F1
&\tiny Support\\ 
\hline
    Background    &   1.00  &    1.00  &    1.00  &     111 \\ \hline
               Cause   &    0.90  &     0.70   &   0.79   &     82 \\ \hline
          Comparison   &    0.00  &    0.00  &    0.00    &    29 \\ \hline
           Condition  &     1.00  &    0.98  &   0.99 &       48 \\ \hline
            Contrast  &    0.99  &    1.00 &     0.99  &     146 \\ \hline
         Elaboration  &     0.75  &    1.00  &    0.86  &     796 \\ \hline
          Enablement  &    0.84   &   1.00  &    0.91  &      46 \\ \hline
          Evaluation  &      0.99  &    1.00 &     0.99  &      80 \\ \hline
         Explanation  &     0.72   &   0.97  &    0.83   &    110 \\ \hline
               Joint  &     0.75  &   0.03  &    0.05  &     212 \\ \hline
        Manner-Means  &     0.00  &    0.00 &     0.00  &      27 \\ \hline
             Summary  &      0.00 &     0.00  &    0.00 &       32 \\ \hline
            Temporal  &    1.00  &    1.00 &     1.00  &       73 \\ \hline
Textual-Organization  &     0.00 &     0.00 &      0.00  &       9 \\ \hline
        Topic-Change   &    0.28  &   1.00  &    0.44   &     13 \\ \hline
       Topic-Comment   &    0.00  &    0.00  &    0.00  &      24 \\ \hline

\textbf{Accuracy}& \multicolumn{4}{|c|} {\textbf{0.81}} \\ \hline
\textbf{Macro-F1} & \textbf{0.58} & \textbf{0.60} & \textbf{0.55} & 1838 \\ 
\hline
\end{tabular}
\vspace{2mm}
\caption{\label{rst-impl-order-ab} Classification report for RST, with the implication order dimension removed.}
\end{table}

Table~\ref{rst-temp-ab} shows the classification report on RST-DT with BERT embeddings and UniDim dimensions as input features, the temporality dimension being removed. 

\begin{table}[H]\tiny
\centering
\begin{tabular}{|l|l|l|l|l|}
\hline  
&\tiny Precision 
&\tiny Recall 
&\tiny F1
&\tiny Support\\ 
\hline
    Background     &  1.00    &  1.00  &    1.00    &   111  \\ \hline
               Cause   &    0.92   &   0.70  &    0.79   &     82 \\ \hline
          Comparison   &    0.00  &    0.00  &    0.00   &     29  \\ \hline
           Condition  &     1.00  &    0.88  &    0.93   &     48 \\ \hline
            Contrast  &     0.99  &    1.00 &     0.99   &    146 \\ \hline
         Elaboration  &     0.75  &    1.00  &    0.86   &    796 \\ \hline
          Enablement  &     0.84  &     1.00  &    0.91  &      46 \\ \hline
          Evaluation  &    0.99  &    1.00  &    0.99 &       80 \\ \hline
         Explanation  &      0.69 &     0.97 &     0.81 &      110 \\ \hline
               Joint  &     1.00  &    0.03  &     0.06  &     212 \\ \hline
        Manner-Means  &     0.00 &     0.00 &     0.00  &      27 \\ \hline
             Summary  &     0.00 &     0.00  &    0.00  &       32 \\ \hline
            Temporal  &    1.00  &   1.00  &    1.00  &      73 \\ \hline
Textual-Organization  &     0.00  &    0.00  &    0.00  &       9 \\ \hline
        Topic-Change  &     0.28  &    1.00  &    0.44  &     13 \\ \hline
       Topic-Comment  &      0.00  &    0.00  &    0.00  &      24 \\ \hline

\textbf{Accuracy}& \multicolumn{4}{|c|} {\textbf{0.80}} \\ \hline
\textbf{Macro-F1} & \textbf{0.59} & \textbf{0.60} & \textbf{0.55} & 1838 \\ 
\hline
\end{tabular}
\vspace{2mm}
\caption{\label{rst-temp-ab} Classification report for RST, with the temporality dimension removed.}
\end{table}

Table~\ref{rst-add-f-ab} shows the classification report on RST-DT with BERT embeddings and UniDim dimensions as input features, the additional dimensions being removed. 

\begin{table}[H]\tiny
\centering
\begin{tabular}{|l|l|l|l|l|}
\hline  
&\tiny Precision 
&\tiny Recall 
&\tiny F1
&\tiny Support\\ 
\hline
     Background   &    0.95  &    1.00  &    0.97    &   111 \\ \hline
               Cause  &     0.90  &    0.70   &   0.79   &     82  \\ \hline
          Comparison  &      0.00  &    0.00  &     0.00  &      29 \\ \hline
           Condition  &     1.00   &   0.81  &    0.90  &      48 \\ \hline
            Contrast  &     0.99  &    1.00  &   0.99  &     146 \\ \hline
         Elaboration  &     0.75 &     1.00  &    0.86 &      796 \\ \hline
          Enablement  &     0.84  &     1.00  &    0.91 &       46 \\ \hline
          Evaluation   &    0.90  &    1.00  &    0.95 &       80 \\ \hline
         Explanation   &    0.71  &    0.97  &    0.82  &     110 \\ \hline
               Joint  &     0.00  &    0.00  &    0.00  &      212 \\ \hline
        Manner-Means  &    0.00  &    0.00  &    0.00 &       27 \\ \hline
             Summary   &    0.00 &     0.00 &     0.00  &      32 \\ \hline
            Temporal   &    1.00  &   1.00  &    1.00  &      73 \\ \hline
Textual-Organization  &     0.00 &     0.00 &     0.00  &        9 \\ \hline
        Topic-Change  &     0.28 &    1.00   &   0.44    &    13  \\ \hline
       Topic-Comment  &     0.00 &     0.00  &    0.00    &    24  \\ \hline

\textbf{Accuracy}& \multicolumn{4}{|c|} {\textbf{0.80}} \\ \hline
\textbf{Macro-F1} & \textbf{0.52} & \textbf{0.59} & \textbf{0.54} & 1838 \\ 
\hline
\end{tabular}
\vspace{2mm}
\caption{\label{rst-add-f-ab} Classification report for RST, with the additional dimensions removed.}
\end{table}

\section{Cross-framework Discourse Relation Classification}\label{transfer-learning}

Table~\ref{body-total-pdtb-dim} shows the classification report of the experiment using total PDTB data, where PDTB relation classification is the source task. 

\begin{table}[h!]\tiny
\centering
\resizebox{0.8\columnwidth}{!}{%
\begin{tabular} {|c| >{\columncolor[RGB]{230, 242, 255}}c|>{\columncolor[RGB]{230, 242, 255}}c|>{\columncolor[RGB]{230, 242, 255}}c| c|c|c|c|}
\hline  
&\tiny $P$
&\tiny $R$
&\tiny $F1$
&\tiny $P_{b.}$
&\tiny $R_{b.}$
&\tiny $F1_{b.}$
&\tiny $C.$\\ 
\hline
Asynchronous & 1.00 & 1.00 & 1.00 & 0.79 & 0.65 & 0.71 & 232\\ \hline
Cause & 1.00 & 1.00 & 1.00 & 0.71 & 0.62 & 0.66 &   538 \\ \hline
Cause+Belief & 1.00 & 1.00 & 1.00 & 0.00 & 0.00 & 0.00 &  13 \\ \hline
Concession & 0.99 & 0.96 & 0.98 & 0.78 & 0.83 & 0.80 & 371 \\ \hline
Condition & 1.00 & 1.00 & 1.00 & 0.92 & 0.87 & 0.90 &  79 \\ \hline
Conjunction & 0.97 & 1.00 & 0.98 &  0.71 & 0.85 & 0.77 & 745 \\ \hline
Contrast & 1.00 & 1.00 & 1.00 & 0.48 & 0.40 & 0.44 &  102 \\ \hline
Disjunction & 1.00 & 1.00 & 1.00 & 0.86 & 0.90 & 0.88 & 20 \\ \hline
Equivalence & 0.00 & 0.00 & 0.00 & 0.36 & 0.16 & 0.22 & 25 \\ \hline
Instantiation & 0.00 & 0.00 & 0.00 & 0.70 & 0.57 & 0.63 & 117 \\ \hline
Level-of-detail & 0.00 & 0.00 & 0.00 &0.48 & 0.53 & 0.50 & 202 \\ \hline
Manner & 0.07 & 0.96 & 0.14 & 0.41 & 0.62 & 0.49 & 26 \\ \hline
Purpose & 1.00 & 0.96 & 0.98 & 0.87 & 0.84 & 0.85 & 118 \\ \hline
Similarity & 0.00 & 0.00 & 0.00 & 0.78 & 0.58 & 0.67 & 12 \\ \hline
Substitution & 0.68 & 0.91 & 0.78 & 0.53 & 0.49 & 0.51 &  35 \\ \hline
Synchronous & 0.90 & 1.00 & 0.95 & 0.74 & 0.64 & 0.68 & 170 \\ \hline
\textbf{Acc.}& \multicolumn{3}{c|} {\textbf{0.86}} & \multicolumn{4}{c|} {0.71 (vs. DISRPT 2021: 0.74)} \\ \hline
\textbf{Macro-F1} & \textbf{0.66} & \textbf{0.74} & \textbf{0.67} & 0.63 & 0.60 & 0.61 & 2805\\ 
\hline
\end{tabular}
}
\vspace{2mm}
\caption{\label{body-total-pdtb-dim} Results of relation classification on total PDTB data. Blue columns show our results and uncolored columns show results of the baseline model.}
\end{table}

Table~\ref{body-rst-transfer-results} shows the classification report of the target task, i.e. RST relation classification. 

\begin{table}[h!]\tiny 
\centering
\resizebox{0.8\columnwidth}{!}{%
\begin{tabular} {|p{0.9cm}| >{\columncolor[RGB]{230, 242, 255}}c|>{\columncolor[RGB]{230, 242, 255}}c|>{\columncolor[RGB]{230, 242, 255}}c| c|c|c|c|}
\hline  
&\tiny $P$
&\tiny $R$
&\tiny $F1$
&\tiny $P_{b.}$
&\tiny $R_{b.}$
&\tiny $F1_{b.}$
&\tiny $C.$\\ 
\hline
 Background & 1.00 & 1.00 & 1.00 & 0.51 & 0.27 & 0.35 & 111\\ \hline
Cause & 0.90 & 0.70 & 0.79 & 0.17 & 0.07 & 0.10 &  82 \\ \hline
Comparison & 0.00 & 0.00 & 0.00 & 0.42 & 0.38 & 0.40 & 29 \\ \hline
Condition & 1.00 & 0.98 & 0.99 & 0.80 & 0.67 & 0.73 & 48 \\ \hline
Contrast & 0.99 & 1.00 & 0.99 &  0.75 & 0.73 & 0.74 & 146 \\ \hline
Elaboration & 0.75 & 1.00 & 0.86 &  0.60 & 0.82 & 0.69 &  796 \\ \hline
Enablement & 0.92 & 1.00 & 0.96 & 0.48 & 0.78 & 0.60 & 46 \\ \hline
Evaluation & 1.00 & 1.00 & 1.00 & 0.00 & 0.00 & 0.00 & 80 \\ \hline
Explanation & 0.72 & 0.97 & 0.83 & 0.40 & 0.15 & 0.22 & 110 \\ \hline
Joint & 0.00 & 0.00 & 0.00 & 0.57 & 0.66 & 0.61 & 212 \\ \hline
Manner-Means & 0.00 & 0.00 & 0.00 & 0.43 & 0.33 & 0.38 & 27 \\ \hline
Summary & 0.00 & 0.00 & 0.00 &  0.00 & 0.00 & 0.00 & 32 \\ \hline
Temporal & 1.00 & 1.00 & 1.00 & 0.53 & 0.36 & 0.43 & 73 \\ \hline
Textual-Organization & 0.00 & 0.00 & 0.00 & 0.00 & 0.00 & 0.00 & 9 \\ \hline
Topic-Change & 0.28 & 1.00 & 0.44 & 0.00 & 0.00 & 0.00 & 13 \\ \hline
Topic-Comment & 0.71 & 0.21 & 0.32 & 0.00 & 0.00 & 0.00 &24 \\ \hline
\textbf{Acc.}& \multicolumn{3}{c|} {\textbf{0.81}} & \multicolumn{4}{c|} {0.58} \\ \hline
\textbf{Macro-F1} & \textbf{0.58} & \textbf{0.62} & \textbf{0.57} & 0.35 & 0.33 & 0.33 & 1838\\  \hline
RST acc & \multicolumn{3}{c|} {0.81} & \multicolumn{4}{c|} {0.63} \\ \hline
RST Macro-F1 & 0.64 & 0.62 & 0.58 & 0.55 & 0.44 & 0.47 & 1838\\ 
\hline
\end{tabular}
}
\vspace{2mm}
\caption{\label{body-rst-transfer-results} Results of the target task. The results of training a model specifically for RST relation classification with our method are shown in blue columns and the uncolored columns show results of the baseline model.}
\end{table}

\end{document}